\documentclass[10pt,twocolumn,letterpaper]{article}

\usepackage{cvpr}
\usepackage{times}
\usepackage{epsfig}
\usepackage{graphicx}
\usepackage{amsmath}
\usepackage{amssymb}
\newcommand{\norm}[1]{\left\lVert#1\right\rVert}

\usepackage{booktabs}
\usepackage{multirow}
\usepackage{subcaption}
\usepackage{algorithmic}
\usepackage[ruled,vlined]{algorithm2e}

\usepackage[pagebackref=true,breaklinks=true,letterpaper=true,colorlinks,bookmarks=false]{hyperref}

\cvprfinalcopy 

\def\cvprPaperID{6926} 

\ifcvprfinal\fi
\begin{document}
\title{Towards Achieving Adversarial Robustness \\by Enforcing Feature Consistency Across Bit Planes}

\author{Sravanti Addepalli\thanks{Equal contribution}~, ~Vivek B.S.\footnotemark[1]~, ~Arya Baburaj, ~Gaurang Sriramanan, ~R.Venkatesh Babu\\
Video Analytics Lab, Department of Computational and Data Sciences\\ Indian Institute of Science, Bangalore, India\\}

\maketitle

\begin{abstract}
   As humans, we inherently perceive images based on their predominant features, and ignore noise embedded within lower bit planes. On the contrary, Deep Neural Networks are known to confidently misclassify images corrupted with meticulously crafted perturbations that are nearly imperceptible to the human eye. In this work, we attempt to address this problem by training networks to form coarse impressions based on the information in higher bit planes, and use the lower bit planes only to refine their prediction. We demonstrate that, by imposing consistency on the representations learned across differently quantized images, the adversarial robustness of networks improves significantly when compared to a normally trained model. Present state-of-the-art defenses against adversarial attacks require the networks to be explicitly trained using adversarial samples that are computationally expensive to generate. While such methods that use adversarial training continue to achieve the best results, this work paves the way towards achieving robustness without having to explicitly train on adversarial samples. The proposed approach is therefore faster, and also closer to the natural learning process in humans.

\end{abstract}

\section{Introduction}
\thispagestyle{empty}

Deep Neural Networks have been used to achieve remarkable performance in many Computer Vision tasks, such as Classification \cite{krizhevsky2012imagenet}, Segmentation \cite{long2015fully} and Object recognition \cite{redmon2016you}. While these networks achieve near-human accuracy on many benchmark datasets, they are far from being as robust as the human visual system. Deep Networks are known to be vulnerable to carefully crafted imperceptible noise known as Adversarial Perturbations \cite{intriguing-iclr-2014}, which could have disastrous implications in critical applications such as autonomous navigation and surveillance systems. The compelling need of securing these systems, coupled with the goal of improving the worst-case robustness of Deep Networks has propelled research in the area of Adversarial Robustness over the last few years. While adversarial training methods \cite{madry-iclr-2018,zhang2019theoretically} have led to significant progress in improving adversarial robustness, these methods are computationally expensive and also non-intuitive when compared to the learning process in humans. 

Humans perceive images based on features of large magnitude and use finer details only to enhance their impressions \cite{sugase1999global, sripati2009representing}. This \textit{background knowledge} of giving higher importance to information present in higher bit planes naturally equips the human visual system to develop resistance towards adversarial perturbations, which are of relatively lower magnitude. On the contrary, these adversarial perturbations can arbitrarily flip the predictions of Deep Networks to completely unrelated classes, suggesting that such \textit{background knowledge} of giving hierarchical importance to different bit planes is missing in these networks. In this work, we propose to equip Deep Networks with such knowledge, and demonstrate that this improves their robustness to adversarial examples.

We propose a novel \textit{Bit Plane Feature Consistency (BPFC)} regularizer, which can significantly improve adversarial robustness of models, without exposure to adversarial samples during training. The proposed method is considerably faster than methods that require multi-step adversarial samples for training \cite{madry-iclr-2018}, and is therefore scalable to large datasets such as ImageNet. Through this work, we hope to pave the path towards training robust Deep Networks without using adversarial samples, similar to the learning process that exists in human beings. 

The organization of this paper is as follows: The subsequent section presents a discussion on the existing literature related to our work. Section-\ref{section:preliminaries} lays out the preliminaries related to notation and threat model. This is followed by details and analysis of our proposed approach in Section-\ref{section:Proposed_method}. We present the experiments performed and an analysis on the results in Section-\ref{section:experiments}, followed by our concluding remarks in Section-\ref{section:conclusions}. 

The code and pretrained models are available at: \\ \url{https://github.com/val-iisc/BPFC}.

\section{Related Works}
\label{Section:Related_works}
\subsection{Adversarial Training methods}
The most popular methods of improving adversarial robustness of Deep Networks involve Adversarial Training (AT), where clean data samples are augmented with adversarial samples during training. Early formulations such as FGSM-AT~\cite{goodfellow2014explaining} proposed training on adversarial samples generated using single-step optimization, that assumes a first-order linear approximation to the loss function. This was later shown to be ineffective against multi-step attacks by Kurakin~\etal~\cite{kurakin2016adversarial}, wherein the effect of gradient masking was identified. Gradient masking, first identified by Papernot~\etal\cite{papernot2017practical}, is the phenomenon where the trained network yields masked gradients, thereby resulting in the generation of weak adversaries, leading to a false sense of robustness. In a wide variety of settings, numerous countermeasures have been developed that can produce strong adversaries to circumvent gradient masking \cite{tramer2017ensemble}.

Madry~\etal\cite{madry-iclr-2018} proposed Projected Gradient Descent (PGD) based training, that employs an iterative procedure for finding strong adversaries that maximise training loss under the given norm constraint. Crucially, PGD trained models are robust against various gradient-based iterative attacks, as well as several variants of non-gradient based attacks. However, the process of PGD Adversarial Training (PGD-AT) is computationally expensive. 

In order to address this, Vivek \etal~\cite{S_2019_CVPR_Workshops} revisited single-step adversarial training, and introduced a regularizer that helps mitigate the effect of gradient masking. This regularizer penalizes the $\ell_2$ distance between logits of images perturbed with FGSM and R-FGSM \cite{tramer2017ensemble} attacks. In the proposed method, we achieve adversarial robustness without using adversarial samples during training, and hence achieve a further reduction in computation time.
 
\subsection{Attempts of Adversary-Free Training}
In this section, we discuss existing training methods that do not utilize adversarial samples during training. Works such as Mixup~\cite{zhang2017mixup} and Manifold-Mixup~\cite{verma2018manifold} propose training methods to learn better feature representations. In Mixup, the network is trained to map a random convex combination of the input data to the corresponding convex combination of their one-hot encodings. This work is extended further in Manifold Mixup, where the network is trained to map a convex combination of intermediate hidden-layers generated by two different data points to the corresponding convex combination of their one-hot encodings. Hence, these methods encourage the network to behave in a linearized manner between input data points, or between hidden-layers deeper in the network. While these methods resulted in improved performance against single-step FGSM attacks, they were susceptible to stronger multi-step attacks.

Another attempt of adversary-free training to achieve robustness utilized input transformations for defense. In the work by Guo~\etal\cite{guo2018countering}, the effect of various input transformations such as bit-depth reduction, JPEG compression, total variation minimisation and image quilting was studied. The robustness from these techniques primarily originated from the non-differentiable pre-processing steps, in order to possibly thwart gradient-based iterative attacks. This method, along with a few others \cite{buckman2018thermometer,ma2018characterizing,s.2018stochastic,xie2018mitigating,song2018pixeldefend}, were broken in the work by Athalye \etal~\cite{athalye2018obfuscated}, where it was identified that obfuscated gradients do not provide reliable security against adversaries. 

Yet another avenue pursued was towards the detection of adversarial samples. Feature Squeezing, proposed by Xu \etal~\cite{xu2017feature}, used transformations such as reduction of color bit depth, spatial smoothing with a median filter and a combination of both, in order to generate a feature-squeezed image from a given input image. By thresholding the $\ell_1$ distance between logits of an input image and its feature-squeezed counterpart, the image was classified to be either adversarial or legitimate in nature. However, in the work by He \etal~\cite{he2017adversarial}, it was shown that an adaptive attacker cognizant of this defense strategy could fool the model by constructing attacks that retain adversarial properties even after feature-squeezing is applied, thereby evading detection.

While we use the concept of quantization to defend against adversarial attacks in this work, we do not introduce any pre-processing blocks that lead to obfuscated or shattered gradients. 
\begin{figure*}
        \begin{subfigure}[b]{0.192\textwidth}
                \includegraphics[width=\linewidth]{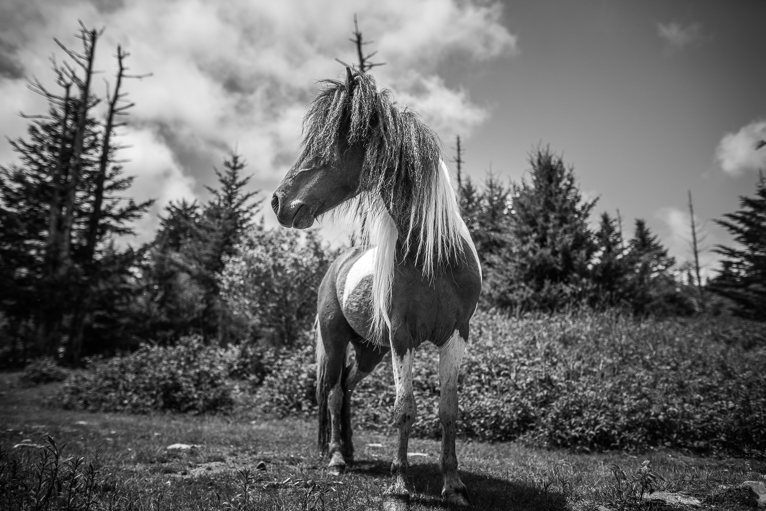}
                \caption{}
        \end{subfigure}%
        \hspace{0.02cm}
        \begin{subfigure}[b]{0.192\textwidth}
                \includegraphics[width=\linewidth]{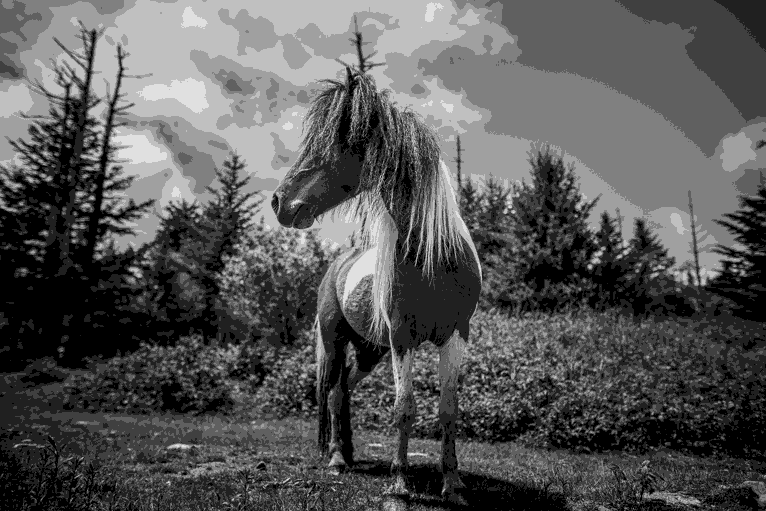}
                \caption{}
        \end{subfigure}%
        \hspace{0.02cm}
        \begin{subfigure}[b]{0.192\textwidth}
                \includegraphics[width=\linewidth]{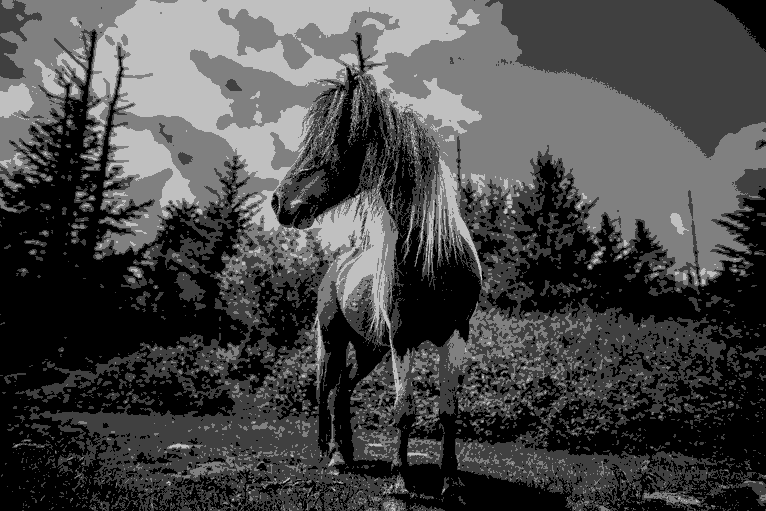}
                \caption{}
        \end{subfigure}%
        \hspace{0.02cm}
        \begin{subfigure}[b]{0.192\textwidth}
                \includegraphics[width=\linewidth]{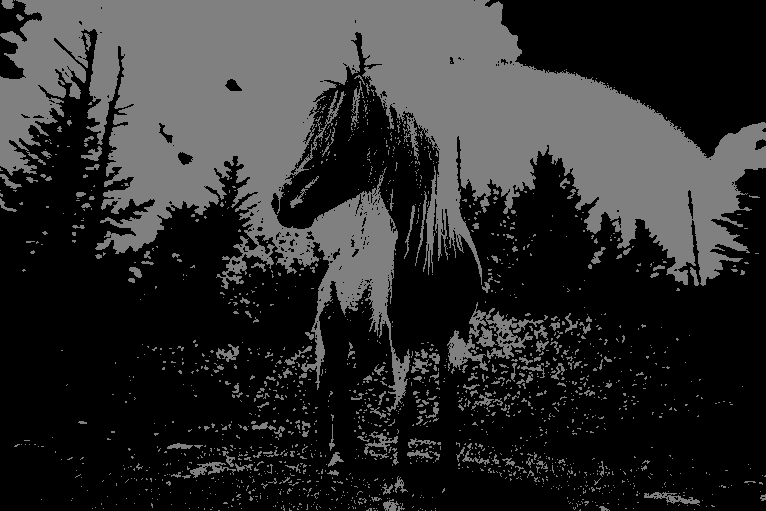}
                \caption{}
        \end{subfigure}%
        \hspace{0.02cm}
        \begin{subfigure}[b]{0.192\textwidth}
                \includegraphics[width=\linewidth]{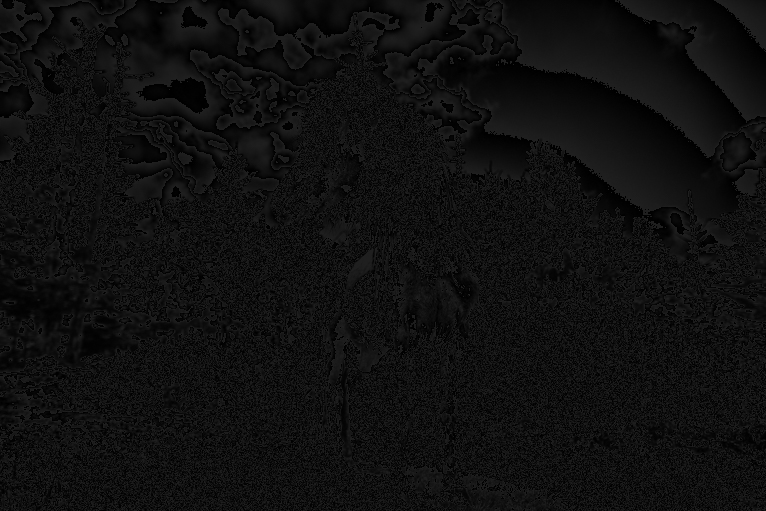}
                \caption{}
        \end{subfigure}
        \vspace{-0.3cm}
        \caption{(a) Original 8-bit image (b) Weighted sum of (higher) bit planes 7, 6 and 5 (c) Weighted sum of (higher) bit planes 7 and 6 (d) Bit plane 7 - Most significant bit plane (e)  Weighted sum of (lower) bit planes 4, 3, 2, 1 and 0.}
        \label{fig:overview_of_bit_quant}
        \vspace{-0.3cm}
\end{figure*}

\section{Preliminaries}
\label{section:preliminaries}
\subsection{Notation}
In this paper, we consider $f(.)$ as the function mapping of a classifier $C$, from an image $x$, to its corresponding softmax output $f(x)$. The predicted class label, which is an argmax over the softmax output, is denoted by $c(x)$. The ground truth label corresponding to $x$ is denoted by $y$. The image is said to be correctly classified when $c(x) = y$. The pre-softmax output of the classifier $C$ is denoted by $g(x)$. We define $\mathcal{A}(x)$ to be the set of all \textit{Adversarial Samples} corresponding to $x$. A specific adversarial sample corresponding to a clean sample $x$ is denoted by $x'$.

\subsection{Threat Model}
\label{subsection:threat_model}
In this paper, we consider the task of improving the worst-case robustness of Deep Networks. The goal of an adversary is to cause an error in the prediction of the classifier. We define an \textit{Adversarial Sample} $x'$, as one that causes the output of the network to be different from the ground truth label $y$. We do not restrict the adversary to flip labels from a specific source class, or to a specific target class. We restrict $x'$ to be in the $\ell_{\infty}$-ball of radius $\varepsilon$ around $x$. The set of \textit{Adversarial Samples} can be formally defined as follows:

\begin{equation}
    \mathcal{A}(x) = \{x':c (x') \neq y, \norm{x - x'}_{\infty} \leq \varepsilon \}
\end{equation}
We therefore impose a constraint that any individual pixel in the image $x$ cannot be perturbed by more than $\varepsilon$. 

Since the goal of this work is to improve worst-case robustness, we do not impose any restrictions on the access to the adversary. We consider that the adversary has complete knowledge of the model architecture, weights and the defense mechanism employed. 

%
\section{Proposed Method}
\label{section:Proposed_method}
In this section, we first present the motivation behind our proposed method, followed by a detailed discussion of the proposed algorithm. We further describe local properties of networks trained using the proposed regularizer, which lead to improved robustness.
%
\subsection{Hierarchical Importance of Bit Planes}
Bit planes of an image are the spatial maps (of the same dimension as the image) corresponding to a given bit position. For an $n$-bit representation of an image, bit plane $n-1$ corresponds to the most significant bit (MSB), and bit plane $0$ corresponds to the least significant bit (LSB). An $n$-bit image can be considered as the sum of $n$ bit planes weighted by their relative importance. The importance of features embedded within lower bit planes is significantly lower than that of features embedded within higher bit planes, both in terms of pixel value, and information content \cite{shan2008image}. Steganography methods \cite{Fridrich:2001:RDL:1232454.1232466} utilize lower bit planes to embed crucial copyright information that needs to be visually imperceptible. However, information content in natural images decreases from the most significant bit (MSB) to the least significant bit (LSB). A weighted sum of the five least significant bit planes of the image in Fig.\ref{fig:overview_of_bit_quant}(a) is shown in Fig.\ref{fig:overview_of_bit_quant}(e), from which it is evident that lower bit planes contribute only to fine details. Fig.\ref{fig:overview_of_bit_quant}(b), (c) and (d) show images ranging from fine to coarse structure, with different levels of quantization. The difference between Fig.\ref{fig:overview_of_bit_quant}(a) and Fig.\ref{fig:overview_of_bit_quant}(b) is Fig.\ref{fig:overview_of_bit_quant}(e). While the addition of Fig.\ref{fig:overview_of_bit_quant}(e) certainly improves the information content, it is not as crucial as the higher bit planes for interpreting the image. 
 
The human visual system is known to give higher importance to global information when compared to fine details \cite{sripati2009representing}. Sugase \etal \cite{sugase1999global} demonstrate that global information is used for coarse classification in early parts of the neural response, while information related to fine details is perceived around 51ms later. This demonstrates a hierarchical classification mechanism, where the response to an image containing both coarse and fine information is aligned with that containing only coarse information.

We take motivation from this aspect of the human visual system, and enforce Deep Networks to maintain consistency across decisions based on features in high bit planes alone (quantized image) and all bit planes (normal image). Such a constraint will ensure that Deep Networks give more importance to high bit planes when compared to lower bit planes, similar to the human visual system. Adversarial examples constrained to the $\ell_\infty$-ball utilize low bit planes to transmit information which is inconsistent with that of higher bit planes. The fact that Deep Networks are susceptible to such adversarial noise demonstrates the weakness of these networks, which emanates from the lack of consistency between predictions corresponding to coarse information and fine details. Therefore, enforcing feature consistency across bit planes results in a significant improvement in adversarial robustness when compared to conventionally trained networks. 

While we use the base-$2$ (binary) representation of an image to illustrate the concept of ignoring low magnitude additive noise, the same can be formulated in terms of any other representation (in any other base) as well. Secondly, low magnitude noise does not always reside in low bit planes. It can overflow to MSBs as well, based on the pixel values in the image. We introduce pre-quantization noise in our proposed approach to mitigate these effects. This is illustrated in the following section, where we explain our proposed method in greater detail.

\subsection{Proposed Training Algorithm}

We present the proposed training method in Algorithm-\ref{alg:BPFC_Training}. Motivated by the need to learn consistent representations for coarse and fine features of an image, we introduce a regularizer that imposes feature consistency between each image and its quantized counterpart. 

\begin{algorithm}[tb]
   \caption{Bit Plane Feature Consistency }
   \label{alg:BPFC_Training}

\begin{algorithmic}
\STATE {\bfseries Input:} Network $f$ with parameters $\theta$, fixed weight $\lambda$, training data $\mathcal{D} = \{(x_i,y_i)\}$ of $n$-bit images, quantiz-\\ation parameter $k$, learning rate $\eta$, minibatch size $M$
\FOR{minibatch $B \subset \mathcal{D} $} 
    
    \STATE Set $L = 0$
        \FOR{$i=1$ {\bfseries to} $M$}
                \STATE $x_{pre} = x_i +  \mathcal{U}(-2^{k-2},2^{k-2})$ ~~~~// \small{Add noise}
                \STATE $x_{q} = x_{pre} - \left(x_{pre} \hspace{0.07in}mod \hspace{0.05in}2^{k} \right)$\hspace{0.05in}~~~~~~~~// \small{Quantization}
                \STATE $x_q = x_q + 2^{k-1}$\hspace{0.74in} ~~~~~~~// \small{Range Shift}
                \STATE $x_q = min(max(x_q,0),2^n -1)$ ~~~~~// \small{Clip}
                \STATE $L = L +  ce(f (x_i) , y_i ) + \lambda \norm{g (x_i) - g(x_q)}^2_2$

        \ENDFOR
        \STATE $\theta = \theta - \frac{1}{M} \cdot \eta \cdot \nabla_{\theta} L\hspace{0.6in}$ ~~~~// \small{SGD update}
    
\ENDFOR
\end{algorithmic}
\end{algorithm}

\subsubsection{Quantization}
\label{CQ}
The steps followed for generating a coarse image are described in this section. The input image $x_i$ is assumed to be represented using $n$-bit quantization. The intensity of pixels is hence assumed to be in the range $[0, 2^n)$. We generate an $n-k+1$ bit image using the quantization process described here. The allowed range of $k$ is between $1$ and $n-1$.
\begin{itemize}
\itemsep0em 
    \item \textbf{Pre-quantization noise}: Initially, uniform noise sampled independently from $~\mathcal{U}(-2^{k-2},2^{k-2})$ is added to each pixel in the image $x_i$, to generate $x_{pre}$.
    \item \textbf{Quantization step}: Next, each pixel is quantized to $n-k$ bits, by setting the last $k$ bits to $0$.
    \item \textbf{Range Shift}: The intensity of all pixels is shifted up by $2^{k-1}$. This shifts the range of quantization error (w.r.t. $x_{pre}$) from $[0, 2^k)$ to $[-2^{k-1}, 2^{k-1})$. 
    \item \textbf{Clip}: Finally, the quantized image is clipped to the original range, $[0, 2^n)$.
\end{itemize}

\begin{figure}[t]
\centering
\fbox{\includegraphics[width=0.95\linewidth]{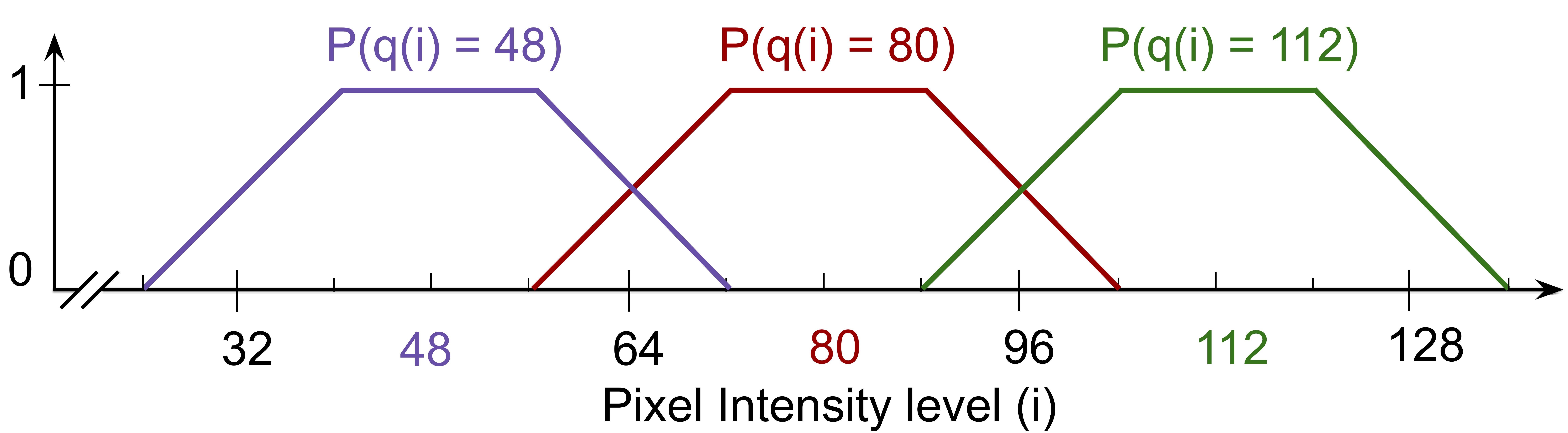}}
\caption{Quantization of a given 8-bit pixel to 3-bits ($n=8,~k=5)$: The probability $P$ of a pixel $i$ being assigned to the quantized values $q(i) =$ $48$, $80$ and $112$ is shown here.}
\label{fig:diag_quant}
\end{figure}
Fig.\ref{fig:diag_quant} illustrates the probability of a pixel $i$ being assigned to different quantization levels using the above method, when $n=8$ and $k=5$. As shown in the figure, addition of pre-quantization noise introduces randomness in the quantized value of each pixel. The probability of being assigned the nearest quantization level is $1$ when the input pixel intensity is close to the quantized value, and decays linearly to $0$ after a fixed distance. This helps mitigate unwanted non-linear effects at the edges of each quantization bin, due to specific pixel intensity values of the original image. 

We do an ablation experiment (in Section-\ref{section:experiments}) called Simple Quantization, where the pre-quantization noise is not added. Simple Quantization can be also viewed as addition of correlated low magnitude (quantization) noise, where the noise pattern depends on local pixel intensity levels. This noise is the difference between the original image, and the image subject to simple quantization. Since the pixel intensities can be assumed to be locally correlated in space, the noise is also correlated locally. The correlated nature of noise differentiates quantization noise from random noise, and also brings it closer to the properties of adversarial perturbations. We also consider an ablation experiment of replacing the quantization step with addition of random noise sampled from a uniform distribution in Section-\ref{section:experiments}. 

While pre-quantization noise disturbs the local correlation properties of the quantization noise for some of the pixels, it is crucial to mitigate the bin edge effects discussed above. We demonstrate through experiments that the proposed solution is better than both the ablation experiments discussed above.

\subsubsection{Bit Plane Feature Consistency Regularizer}
The loss function used for training is shown below:
\begin{equation} \label{eq_loss}
    L = \dfrac{1}{M}\sum_{i=1}^{M} ce(f(x_i),y_i) + \lambda {\norm{g(x_i) - g(q(x_i))}}^2_2
\end{equation}

For a given image $x_i$, the first term of Eq.~(\ref{eq_loss}) is the cross-entropy ($ce$) loss obtained from the softmax output of the network  $f(x_i)$, and the corresponding ground truth label $y_i$. The second term is the squared $\ell_2$ distance between the pre-softmax activation of the image $x_i$, and that of the corresponding quantized image $q(x_i)$ (generated using the process described in Section-\ref{CQ}). We call this squared $\ell_2$ loss term as the \textit{Bit Plane Feature Consistency} (\textit{BPFC}) regularizer, as it ensures that the network learns consistent feature representations across the original image as well as the coarse quantized image. The loss for each minibatch of size $M$ is an average over all samples in the minibatch.

The cross-entropy term on original images ensures that a combination of coarse and fine features is used to learn the overall function mapping $g(.)$. This helps preserve the accuracy on clean images, while the \textit{BPFC} regularizer helps improve the adversarial robustness of the model.

\subsection{Local Properties of BPFC Trained Networks}
\label{subsection:local_properties}
In this section, we examine local properties of the function $g(.)$ learned using the proposed \textit{BPFC} regularizer.

Let $x_i$ denote an $n$-bit image sampled from the data distribution $\mathbb{P}_D$ with pixel intensities in the range $[0,2^n)$, and let $q(x_i)$ denote a quantized image corresponding to $x_i$. We assume that $q(x_i)$ is not identically equal to $x_i$. For a fixed value of $\lambda$, let $\Theta_{g(\lambda)}$ denote the set of parameters corresponding to a family of functions that lead to the cross-entropy term in Eq.~(\ref{eq_loss}) being below a certain threshold. Minimization of \textit{BPFC} loss among the family of functions parameterized by $\Theta_{g(\lambda)}$ is shown in Eq.~(\ref{lipschitz_pf1}):
\begin{flalign}
\label{lipschitz_pf1}
    &~~~~~~~~~\underset{\theta_{g} \in \Theta_{g(\lambda)}}{\operatorname{min}} \mathbb{E}_{x_i\sim\mathbb{P}_D}~\mathbb{E}_{q(x_i)}\norm{g(x_i) - g(q(x_i))}^2_2 &
\end{flalign}
\vspace{-0.6cm}
\begin{flalign}
\label{lipschitz_pf2}
   &~~~~~~~~~\underset{\theta_{g} \in \Theta_{g(\lambda)}}{\operatorname{min}} {\mathbb{E}_{x_i\sim\mathbb{P}_D}~\mathbb{E}_{q(x_i)}} \dfrac{{\norm{g(x_i) - g(q(x_i))}}^2_2}{{\norm{x_i - q(x_i)}}^2_2}&
\end{flalign}

The expression in Eq.~(\ref{lipschitz_pf1}) can be lower bounded by the expression in Eq.~(\ref{lipschitz_pf2}), which is equivalent to minimizing the local Lipschitz constant of the network at each sample $x_i$. The denominator of the objective function in Eq.~(\ref{lipschitz_pf2}) is the $\ell_2$-norm between each image and its quantized counterpart, and is thus independent of $\theta_g$. Therefore, minimization of \textit{BPFC} loss in Eq.~(\ref{lipschitz_pf1}) can be viewed as minimization of the local Lipschitz constant at each sample $x_i$, weighted by an $\ell_2$-norm of its deviation from the quantized image. An expectation of this $\ell_2$-norm term over all $q(x_i)$ (with differently sampled pre-quantization noise) converges to a constant value for all samples, thereby establishing an equivalence between the minimization of \textit{BPFC} loss and the minimization of local Lipschitz constant of the network.

Hence, imposing \textit{BPFC} regularizer encourages the network to be locally Lipschitz continuous with a reduced Lipschitz constant. While the \textit{BPFC} regularizer imposes local smoothness, the cross-entropy term in Eq.~(\ref{eq_loss}) requires $g(.)$ to be a complex mapping for better accuracy on clean images. The final selection of $\theta_g$ would depend on $\lambda$, which is typically selected based on the amount by which clean accuracy can be traded-off for adversarial accuracy \cite{zhang2019theoretically, tsipras2018robustness}. During the initial epochs of training, the function learned is relatively smooth. Hence, we start with a low value of $\lambda$ and step it up during training. 

Therefore, the \textit{BPFC} formulation leads to functions with improved local properties, which is closely related to adversarial robustness as explained by Szegedy \etal \cite{intriguing-iclr-2014}.

\section{Experiments and Analysis}
\label{section:experiments}

In this section, we discuss the experiments done to verify the robustness of our proposed approach. We first describe the datasets used and details on the training methodology in Section-\ref{subsection:experiments_Preliminaries}, followed by an overview of the experiments conducted in Section-\ref{subsection:experiments_overview}. We further present details on each experiment and our analysis on the results in Sections-\ref{subsection:experiments_wb} to \ref{subsection:experiments_adaptive}. We follow the guidelines laid out by Athalye \etal \cite{athalye2018obfuscated} and Carlini \etal \cite{carlini2019evaluating} to ascertain the validity of our claim on the achieved robustness. 

\subsection{Preliminaries}
\label{subsection:experiments_Preliminaries}

We use the benchmark datasets, CIFAR-$10$ \cite{cifar_10_dataset}, Fashion-MNIST (F-MNIST) \cite{fashion_mnist} and MNIST \cite{lecun1998mnist} for validating our proposed approach. CIFAR-$10$ is a ten-class dataset with RGB images of dimension $32 \times 32$. The number of images in the training set and test set are $50,000$ and $10,000$ respectively. These images are equally distributed across all classes. We set aside $10,000$ images from the training set as the validation set. Fashion-MNIST and MNIST are ten-class datasets with gray-scale images of dimension $28 \times 28$. The datasets are composed of $60,000$ training samples and $10,000$ test samples each. We further split each of the training datasets into $50,000$ training samples and $10,000$ validation samples. 

We use ResNet-18 \cite{he2016deep} architecture for CIFAR-$10$, and a modified LeNet (M-LeNet) \cite{lecun1998gradient} architecture with two additional convolutional layers (details in Table-\ref{table:mnsit_architecure} of the Supplementary) for MNIST and Fashion-MNIST. We train CIFAR-$10$ models for $100$ epochs, MNIST and Fashion-MNIST models for 50 epochs each. The minibatch size is set to $128$ for CIFAR-$10$ and $64$ for Fashion-MNIST and MNIST. We use SGD optimizer with momentum of $0.9$ and weight decay of $5\mathrm{e}$-${4}$. We use an initial learning rate of $0.1$ for CIFAR-$10$, and $0.01$ for MNIST and Fashion-MNIST. We reduce the initial learning rate by a factor of $5$, three times during the training process. We use early stopping based on I-FGSM \cite{kurakin2016physical} accuracy on the validation split in the last 20 epochs for CIFAR-$10$, and last 30 epochs for MNIST and Fashion-MNIST. 

The hyperparameters to be selected for training are: $k$, which is the number of bits to be eliminated during the quantization step in Section-\ref{CQ}, and $\lambda$, which is the weighting factor for \textit{BPFC} loss in Eq.~(\ref{eq_loss}). We set $k$ to $5$ for CIFAR-$10$, $6$ for Fashion-MNIST and $7$ for MNIST. The value of $k$ can be selected based on the $\varepsilon$ value of the attack to be defended. $\lambda$ is selected to achieve the desired trade-off between accuracy on clean and adversarial samples (details in Section-\ref{subsection:suppl_impact_of_hyperparameters} of the Supplementary). As explained in Section-\ref{subsection:local_properties}, we start with a low value of $\lambda$ and step it up over epochs. This helps achieve better accuracy on clean samples. For CIFAR-$10$, we start with $\lambda$ of 1 and step it up by a factor of $9$ every 25 epochs. Since the clean accuracy on Fashion-MNIST and MNIST datasets increases within very few epochs, we use high $\lambda$ values from the beginning (without a step-up factor). We thus use a $\lambda$ value of $30$ for MNIST and $25$ for Fashion-MNIST. 

\subsection{Overview of Experiments}
\label{subsection:experiments_overview}
We compare the proposed approach with Normal Training (NT), FGSM-AT \cite{goodfellow2014explaining}, PGD-AT \cite{madry-iclr-2018} and Regularized Single-Step Adversarial Training (RSS-AT) \cite{S_2019_CVPR_Workshops} across all three datasets. We report results on single-step (FGSM) and multi-step (I-FGSM, PGD) attacks, epsilon-bounded and unbounded (DeepFool \cite{moosavi2016deepfool}, Carlini-Wagner (C\&W) \cite{carlini2017towards}) attacks, untargeted and targeted attacks, and gradient-free attacks (random attacks, SPSA \cite{uesato2018adversarial}). We consider attacks in white-box and black-box settings. We also consider adaptive attacks that are specific to the defense mechanism used. 

As explained in Section-\ref{subsection:threat_model}, we restrict the adversary to be in the $\ell_\infty$-ball of radius $\varepsilon$ around each data point. We refer to the work by Madry \etal \cite{madry-iclr-2018} for attack parameters and number of iterations for PGD attack. For an image with pixel intensities in the range $[0, 1]$, we consider an $\varepsilon$ value of $8/255$ for CIFAR-$10$, $0.3$ for MNIST and $0.1$ for Fashion-MNIST. We consider $\varepsilon_{step}$ to be $2/255$, $0.01$ and $0.01$ for CIFAR-$10$, MNIST and Fashion-MNIST respectively. These restrictions do not apply to the unbounded attacks, DeepFool and C\&W. 

We present our experiments, results and analysis for each attack in the following subsections. 

\begin{table}[]
\caption{\textbf{CIFAR-10}: Recognition accuracy (\%) of models in a white-box attack setting.}
\label{table:cifar10_whitebox}
\vspace{-0.0cm}
\setlength\tabcolsep{3pt}
\resizebox{1.0\linewidth}{!}{
\begin{tabular}{l|ccc|ccc}
\toprule
\multirow{2}{*}{\textbf{\begin{tabular}[c]{@{}l@{}}Training method\end{tabular}}}     & \multirow{2}{*}{\textbf{\begin{tabular}[c]{@{}l@{}}Clean\end{tabular}}} & \multirow{2}{*}{\textbf{\begin{tabular}[c]{@{}l@{}}FGSM\end{tabular}}} & \textbf{IFGSM} & \multicolumn{3}{c}{\textbf{PGD (n-steps)}}            \\
                     &                &               & \textbf{7 steps}        & \textbf{~~~~7~~~~} & \textbf{~~20~~} & \textbf{1000}                      \\\midrule
FGSM-AT              & 92.9 & 96.9 & 0.8  & 0.4  & 0.0  & 0.0  \\
RSS-AT               & 82.3 & 55.0 & 50.9 & 50.0 & 46.2 & 45.8 \\
PGD-AT               & 82.7 & 54.6 & 51.2 & 50.4 & 47.4 & 47.0 \\ \midrule
NT               & 92.3 & 16.0 & 0.0  & 0.0  & 0.0  & 0.0  \\
Mixup                & 90.3 & 27.4 & 1.6  & 0.6  & 0.1  & 0.0  \\
BPFC (\textbf{Ours}) & 82.4 & 50.1 & 44.1 & 41.7 & 35.7 & 34.4 \\ \midrule
\multicolumn{7}{c}{~~~~~~~~~~~~~~~~~~~~~~~~~~~~~~Ablations of the proposed approach (BPFC)} \\\midrule
A1: Simple quant & 82.6 & 49.2 & 41.4 & 38.8 & 31.6 & 30.1 \\ 
A2: Uniform noise   & 82.6 & 48.7 & 42.3 & 40.0 & 33.3 & 31.9 \\
A3: $\ell_1$ norm \footnotemark[1]       & 92.1 & 68.3 & 60.8 & 57.1 & 46.8 & 35.9
\\ \bottomrule
\end{tabular}
}
\vspace{-0.2cm}
\end{table}
\footnotetext[1]{A3: The $500$-step worst case PGD accuracy goes down from $37.5\%$ to $24.8\%$ with $100$ random restarts (over $1000$ test samples)}


\subsection{Performance against White-box Attacks}
\label{subsection:experiments_wb}
As explained in Section-\ref{subsection:threat_model}, we consider that the adversary has access to the network architecture and weights. In this scenario, white-box attacks are expected to be stronger than black-box attacks (unless the model merely appears to be robust due to gradient masking). In this section, we consider the following types of white-box attacks: untargeted and targeted $\varepsilon$-bounded attacks, and unbounded attacks.
\begin{table}[]
\centering
\small
\caption{\textbf{Computational complexity} measured in terms of absolute training time per epoch (seconds) and ratio w.r.t. the proposed method (BPFC). This experiment is run on a single Nvidia Titan-X GPU card.}
\label{table:computational_complexity}
\vspace{-0.0cm}
\setlength\tabcolsep{3pt}
\resizebox{1.0\linewidth}{!}{
\begin{tabular}{@{}l|cc|cc|cc@{}}
\toprule
\multirow{2}{*}{\textbf{\begin{tabular}[c]{@{}l@{}}Training \\ method\end{tabular}}}   & \multicolumn{2}{c|}{\textbf{CIFAR-10}} & \multicolumn{2}{c|}{\textbf{F-MNIST}} & \multicolumn{2}{c}{\textbf{MNIST}} \\ \cline{2-7} 
\multicolumn{1}{c|}{}                        & seconds             & ~ratio~~            & ~seconds            & ~ratio~~            & ~seconds           & ~ratio~~          \\ \midrule
RSS-AT                                       & 127.2               & 1.8              & 23.8               & 2.0              & 24.1              & 1.7            \\
PGD-AT                                       & 257.8               & 3.7              & 199.6              & 16.9             & 199.2             & 14.2           \\ \midrule
NT                                           & 39.6                & 0.6              & 9.3                & 0.8              & 8.9               & 0.6            \\
BPFC (\textbf{Ours})~~                                  & 69.4                & 1.0              & 11.8               & 1.0              & 14.0              & 1.0 \\
\bottomrule
\end{tabular}}
\vspace{-0.3cm}
\end{table}

\subsubsection{Bounded Attacks: Untargeted}
\label{subsection:experiments_wb_bounded}
The results of various single-step and multi-step white-box attacks for CIFAR-$10$ dataset are presented in Table-\ref{table:cifar10_whitebox}. FGSM-AT achieves the best robustness to single-step attacks. However, it is not robust to multi-step attacks as explained by Kurakin \etal \cite{kurakin2016adversarial}. PGD-AT and RSS-AT show the best accuracy of around $45\%$ for $1000$-step PGD attack. Mixup \cite{zhang2017mixup} does not use Adversarial Training and achieves an improvement over Normal Training (NT) in robustness towards FGSM attack. However, it is not robust to PGD attacks. The proposed method achieves a significant improvement over Normal Training and Mixup in robustness to both single-step and multi-step attacks, despite not being exposed to adversarial samples during training. As shown in Table-\ref{table:computational_complexity}, the proposed method is faster than methods that are robust to multi-step attacks (PGD-AT and RSS-AT). 
 
As explained in Section-\ref{CQ}, we consider ablation experiments of Simple Quantization (A$1$) and addition of Uniform Noise (A$2$). The proposed method (\textit{BPFC}) achieves an improvement over these two baselines, indicating the significance of the proposed formulation. Adding uniform random noise in the range $(-8/255, 8/255)$ produces an effect similar to that of quantization by reducing the importance given to LSBs for the classification task. Hence, we see comparable results even for this ablation experiment. 

We also consider an ablation experiment of using $\ell_1$-norm instead of $\ell_2$-norm in Eq.~(\ref{eq_loss}). While the results using $\ell_1$-norm (Table-\ref{table:cifar10_whitebox}) show an improvement over the proposed method, the $500$-step worst-case PGD accuracy goes down from $37.5\%$ to $24.8\%$ with $100$ random restarts (over $1000$ test samples represented equally across all classes), indicating that it achieves robustness due to gradient masking. For the proposed approach, the PGD accuracy with $50$ steps ($34.68\%$) is similar to that with $1000$ steps ($34.44\%$). Hence, we check $50$-step PGD accuracy with $1000$ random restarts (for $1000$ test samples) and find that the drop in accuracy over multiple random restarts is negligible. The accuracy drops from $35.6\%$ to $34.9\%$ over $1000$ random restarts, verifying that the robustness in the proposed approach is not due to gradient masking. 

Table-\ref{table:all_dataset_whitebox} shows the consolidated white-box results for all datasets. The proposed method has significantly better robustness to multi-step attacks when compared to methods that do not use Adversarial training (NT and Mixup). We also achieve results comparable to PGD-AT and RSS-AT, while being significantly faster. Detailed results with Fashion-MNIST and MNIST datasets are reported in Tables-\ref{table:fmnist_whitebox} and \ref{table:mnist_whitebox} of the Supplementary.

\begin{table}[]
\caption{\textbf{White-box setting:} Recognition accuracy (\%) of different models on clean samples and adversarial samples generated using PGD-$1000$ step attack.}
\label{table:all_dataset_whitebox}
\vspace{-0.2cm}
\setlength\tabcolsep{3pt}
\resizebox{1.0\linewidth}{!}{
\begin{tabular}{@{}l|cc|cc|cc@{}}
\toprule
\multirow{2}{*}{\textbf{\begin{tabular}[c]{@{}l@{}}Training \\ method\end{tabular}}} & \multicolumn{2}{|c|}{\textbf{CIFAR-10}}                             & \multicolumn{2}{|c|}{\textbf{F-MNIST}}                        & \multicolumn{2}{c}{\textbf{MNIST}}                                \\ \cmidrule(l){2-7} 
\multicolumn{1}{l}{}                  & \multicolumn{1}{|l}{~~Clean~~} & \multicolumn{1}{l|}{~PGD~~} & \multicolumn{1}{|l}{~~Clean~~} & \multicolumn{1}{l}{~PGD~~} & \multicolumn{1}{|l}{~~Clean~} & \multicolumn{1}{l}{~PGD~~} \\ \midrule
FGSM-AT                               & 92.9                      & 0.0                          & 93.1                      & 15.1                         & 99.4                      & 3.7                          \\
RSS-AT                                & 82.3                      & 45.8                         & 87.7                      & 71.8                         & 99.0                      & 90.4                         \\
PGD-AT                                & 82.7                      & 47.0                         & 87.5                      & 79.1                         & 99.3                      & 94.1                         \\ \midrule
NT                                & 92.3                      & 0.0                          & 92.0                      & 0.3                          & 99.2                      & 0.0                          \\
Mixup                                 & 90.3                      & 0.0                          & 91.0                      & 0.0                          & 99.4                      & 0.0                          \\
BPFC (\textbf{Ours})~~~~~                       & 82.4                      & 34.4                         & 87.2                      & 67.7                         & 99.1                      & 85.7                         \\ \bottomrule
\end{tabular}
}
\end{table}

\subsubsection{Bounded attacks: Targeted}
\label{subsection:experiments_wb_targeted}
We evaluate the robustness of \textit{BPFC} trained models against targeted attacks of two types, as discussed in this section. In the first attack (Least Likely target), we set the target class to be the least likely predicted class of a given image. In the second variant (Random target), we assign random targets to each image. We use $1000$-step PGD attacks for both these evaluations, and compare the robustness against an untargeted PGD attack in Table-\ref{table:non_targeted_attack}. As expected, model trained using the proposed approach is more robust to targeted attacks, when compared to an untargeted attack. 
\begin{table}[]
\centering
\caption{Recognition accuracy $(\%)$ of the proposed method (BPFC) on different $1000$-step PGD attacks.}
\label{table:non_targeted_attack}
\vspace{-0.2cm}
\setlength\tabcolsep{3pt}
\resizebox{1.0\linewidth}{!}{
\begin{tabular}{@{}l|c|c|c@{}}
\toprule
\textbf{Attack} & \multicolumn{1}{|c|}{\textbf{CIFAR-10}} & \multicolumn{1}{c|}{\textbf{F-MNIST}} & \multicolumn{1}{c}{\textbf{MNIST}} \\ \midrule
Untargeted             & 34.4                         & 67.7                              & 85.7                      \\
Targeted (Least Likely target)          & 65.2                         & 85.5                              & 95.6                      \\
Targeted (Random target)        & 63.1                         & 83.5                              & 94.8                      \\ \bottomrule
\end{tabular}
}
\end{table}

\begin{table}[]
\caption{\textbf{DeepFool and C\&W attacks (CIFAR-10):} Average $\ell_2$ norm of the generated adversarial perturbations is reported. Higher $\ell_2$ norm implies better robustness. Fooling rate (FR) represents percentage of test set samples that are misclassified.}
\label{table:deepfool_cw_attack}
\vspace{-0.2cm}
\centering
\setlength\tabcolsep{3pt}
\resizebox{1.0\linewidth}{!}{
\begin{tabular}{@{}l|cc|cc@{}}
\toprule
\multirow{2}{*}{\textbf{\begin{tabular}[c]{@{}l@{}}Training \\ method\end{tabular}}} & \multicolumn{2}{|c}{\textbf{DeepFool}}                                   & \multicolumn{2}{|c}{\textbf{C\&W}}                                       \\ \cmidrule(l){2-5} 
                  & \multicolumn{1}{|c}{~~~FR $(\%)$~~~} & \multicolumn{1}{c|}{~~~Mean $\ell_2$~~~} & \multicolumn{1}{c}{~~~FR $(\%)$~~~} & \multicolumn{1}{c}{~~Mean $\ell_2$~~} \\ \midrule
FGSM-AT           & 95.12                            & 0.306                       & 100                              & 0.078                       \\
PGD-AT            & 90.78                            & 1.098                       & 100                              & 0.697                       \\
RSS-AT            & 89.75                            & 1.362                       & 100                              & 0.745                       \\ \midrule
NT                      & 94.66                            & 0.176                       & 100                              & 0.108                       \\
Mixup                   & 93.37                            & 0.168                       & 100                              & 0.104                       \\
BPFC (\textbf{Ours})~~~~    & 89.51                            & 2.755                       & 100                              & 0.804                       \\ \bottomrule
\end{tabular}
}
\vspace{-0.2cm}
\end{table}

\vspace{-0.2cm}
\subsubsection{Unbounded Attacks}
\label{subsection:experiments_wb_unbounded}
We evaluate robustness of \textit{BPFC} trained models to the unbounded attacks, DeepFool and Carlini-Wagner (C\&W). The goal here is to find the lowest $\ell_2$-norm bound on perturbations that can result in $100\%$ fooling rate for all samples. We select the following hyperparameters for C\&W attack: search steps =  $9$, max iterations = $200$, learning rate = $0.01$. For DeepFool attack, we set the number of steps as $100$. With these settings, we achieve $100\%$ fooling rate for C\&W attack with all the training methods. DeepFool does not achieve $100\%$ fooling rate for any of the methods. However, these results are consistent with the results reported in literature \cite{rony2019decoupling}. The performance of these models is measured in terms of average $\ell_2$-norm of the generated perturbations. A higher value on the bound implies that the model has better robustness. The results on CIFAR-$10$ are presented in Table-\ref{table:deepfool_cw_attack}. It can be observed that the \textit{BPFC} trained model is more robust to C\&W attack when compared to all other methods, including PGD-AT. The DeepFool results of the proposed method can be compared directly only with PGD-AT and RSS-AT, as they achieve similar fooling rates. The proposed method achieves significantly improved robustness when compared to both these methods. Results on Fashion-MNIST and MNIST datasets are presented in Section-\ref{subsection:suppl_unbounded_attacks} of the Supplementary.


\begin{table}[]
\caption{\textbf{Black-box setting:} Recognition accuracy (\%) of different models on FGSM black-box adversaries. Columns represent the source model used for generating the attack.}
\label{table:all_blackbox}
\vspace{-0.2cm}
\setlength\tabcolsep{3pt}
\resizebox{1.0\linewidth}{!}{
\begin{tabular}{@{}l|cc|cc|cc@{}}
\toprule
                \multirow{2}{*}{\textbf{\begin{tabular}[c]{@{}l@{}}Training \\ method\end{tabular}}}& \multicolumn{2}{|c|}{\textbf{CIFAR-10}} & \multicolumn{2}{|c|}{\textbf{Fashion-MNIST}} & \multicolumn{2}{c}{\textbf{MNIST}} \\ \cmidrule(l){2-7} 
                & VGG19      & ResNet18      & Net-A           & M-LeNet       & Net-A           & M-LeNet      \\
                \midrule
FGSM-AT         & 78.67       & 77.58          & 94.36       & 90.76        & 87.99       & 85.68       \\
RSS-AT          & 79.80        & 79.99          & 84.99       & 84.16        & 95.28       & 95.19       \\
PGD-AT          & 80.24       & 80.53          & 84.99       & 85.68        & 95.75       & 95.36       \\\midrule
NT              & 36.11       & 15.97          & 34.71       & 16.67        & 29.94       & 16.60       \\
Mixup           & 42.67       & 43.41          & 54.65       & 66.31        & 58.47       & 69.46       \\
BPFC (\textbf{Ours}) & 78.92       & 78.98          & 81.38       & 83.46        & 94.17       & 94.56       \\ \bottomrule
\end{tabular}
}
\vspace{-0.2cm}
\end{table}

\subsection{Performance against Black-box Attacks}
\label{subsection:experiments_bb}
We report accuracy against FGSM black-box attacks in Table-\ref{table:all_blackbox}. We consider two source models for generation of black-box attacks on each dataset; the first is a model with a different architecture, and the second is a model with the same architecture as that of the target model. In both cases, and across all datasets, black-box accuracies are significantly better with the proposed approach when compared to other non-adversarial training methods (NT and Mixup). Further, our results are comparable with those of adversarial training methods. Results on multi-step black-box attacks are presented in Section-\ref{subsection:suppl_blackbox_attacks} of the Supplementary. 
\subsection{Performance against Gradient-free Attacks}
\label{subsection:experiments_gradient_free}
We check the robustness of our proposed approach against the following gradient-free attacks on the CIFAR-$10$ dataset, to ensure that there is no gradient masking: attack with random noise \cite{carlini2019evaluating} and SPSA attack \cite{uesato2018adversarial}. 

For the attack with random noise, we consider a random sample of $1000$ images from the test set of CIFAR-$10$, such that all ten classes are equally represented. We randomly select $10^5$ samples from an $\ell_\infty$-ball of radius $\varepsilon$ around each data point (each pixel is an \textit{i.i.d.} sample from a Uniform distribution) and compute the accuracy of these samples. We find that the accuracy on these random samples is $79.76\%$, which is slightly less than the accuracy on clean samples ($82.4\%$). We run another experiment to verify that every image that is robust to PGD attack is also robust to random noise. We run PGD attack for $50$ steps and $100$ random restarts and identify the images that are robust to the attack. We attack these images with $10^5$ random noise-perturbations each, and find that we achieve the expected accuracy of $100\%$. Hence, we conclude that the attack with random noise is not stronger than a gradient-based attack. 

The SPSA attack \cite{uesato2018adversarial} is a gradient-free attack that computes a numerical approximation of the gradient along multiple random directions and approximates the final gradient to be an average over these gradients. The attack becomes stronger as more directions are used. We use the following hyperparameters to generate the attack: $\delta$ $= 0.01$, learning rate $=0.01$, batch size $= 128$ and iterations $= 5$. We get an accuracy of $70.5\%$ against the SPSA attack using the proposed approach. For the same attack, the accuracy of a PGD trained model is $70.8\%$. 

Therefore, we verify that gradient-based attacks are stronger than gradient-free attacks, thereby confirming the absence of gradient masking. 

\subsection{Performance against Adaptive Attacks}
\label{subsection:experiments_adaptive}
In this section, we consider methods that utilize knowledge of the defense mechanism in creating stronger attacks. We explore maximizing loss functions that are different from the standard cross-entropy loss for generating adversarial samples. We consider the CIFAR-$10$ dataset for this experiment. Maximizing the same loss that is used for training, with the same hyperparameters, gives a slightly lower accuracy ($34.52\%$) when compared to PGD ($34.68\%$) for a $50$-step attack. However, this difference is not statistically significant, and it may be due to the random nature of the PGD attack. The worst accuracy across different hyperparameters in the loss function is $34.41\%$. 

We also explore adding another term to the loss that is maximized during a PGD attack. In addition to maximizing the training loss, we minimize the magnitude of $k$ ($=5$) LSBs in the generated samples. This would encourage the adversaries to have low magnitude LSBs, which could possibly be the samples where the defense was less effective. However, even with this change, we get the same accuracy as that of a standard PGD attack. 

Therefore, the adaptive attacks are only as strong as a PGD attack. We include more details on adaptive attacks in Section-\ref{subsection:suppl_adaptive_attacks} of the Supplementary. 

\subsection{Basic Sanity Checks to Verify Robustness}
\label{subsection:sanity_checks}
In this section, we present results on the basic sanity checks listed by Athalye \etal \cite{athalye2018obfuscated} to ensure that the model's robustness is not due to gradient masking. 
\begin{itemize}
\itemsep0em 
    \item Results in Table-\ref{table:cifar10_whitebox} illustrate that iterative attacks (PGD and I-FGSM) are stronger than an FGSM attack.
    \item White-box attacks are stronger than black-box attacks based on results in Tables-\ref{table:all_dataset_whitebox} and \ref{table:all_blackbox}.
    \item We note that unbounded attacks reach $100\%$ success rate, and increasing distortion bound increases the success rate of the attack (Fig.\ref{fig:acc_vs_eps} of the Supplementary).
    \item As discussed in Section-\ref{subsection:experiments_gradient_free}, gradient-based attacks are stronger than gradient-free attacks. 
    \item We note that cross-entropy loss on FGSM samples increases monotonically with an increase in perturbation size. (Fig.\ref{fig:loss_vs_eps} of the Supplementary). 
\end{itemize}

\subsection{Scalability of the Proposed Method to ImageNet}

We present results on ImageNet \cite{imagenet_cvpr09}, which is a $1000$-class dataset with $1.2$ million images in the training set and $50,000$ images in the validation set. The accuracy on a targeted PGD $20$-step attack is $32.91\%$ with the proposed approach and $43.43\%$ with a PGD-AT model \cite{robustness}. The trend in robustness when compared to PGD-AT is similar to that of CIFAR-$10$ (Table-\ref{table:cifar10_whitebox}), thereby demonstrating the scalability of the proposed approach to large scale datasets. We present detailed results in Section-\ref{subsection:suppl_imagenet} of the Supplementary.

\section{Conclusions}
\label{section:conclusions}
 We have proposed a novel Bit Plane Feature Consistency (\textit{BPFC}) regularizer, which improves the adversarial robustness of models using a normal training regime. Results obtained using the proposed regularizer are significantly better than existing non-adversarial training methods, and are also comparable to adversarial training methods. Since the proposed method does not utilize adversarial samples, it is faster than adversarial training methods. We demonstrate through extensive experiments that the robustness achieved is indeed not due to gradient masking. Motivated by human vision, the proposed regularizer leads to improved local properties, which results in better adversarial robustness. We hope this work would lead to further improvements on the front of non-adversarial training methods to achieve adversarial robustness in Deep Networks. 

\section{Acknowledgements}
This work was supported by RBCCPS, IISc and Uchhatar Avishkar Yojana (UAY) project (IISC\_10), MHRD, Govt. of India. We would like to extend our gratitude to all the reviewers for their valuable suggestions.

{\small
\bibliographystyle{ieee_fullname}
\bibliography{ms}
}



\clearpage







\twocolumn[
  \begin{@twocolumnfalse}
\begin{center}
\textbf{\Large Supplementary material}
\vspace{1cm}
\end{center}
 \end{@twocolumnfalse}
  ]

\setcounter{equation}{0}
\setcounter{figure}{0}
\setcounter{table}{0}
\setcounter{section}{0}
\makeatletter
\renewcommand{\theequation}{S\arabic{equation}}
\renewcommand{\thefigure}{S\arabic{figure}}
\renewcommand{\thetable}{S\arabic{table}}
\renewcommand{\thesection}{S\arabic{section}}



\section{Details on Architecture and Training}
In this section, we present details related to the architecture of models used and the impact of change in hyperparameters.
\subsection{Architecture details for Fashion-MNIST and MNIST}
We use a modified LeNet architecture for all our experiments on Fashion-MNIST and MNIST datasets. This architecture has two additional convolutional layers when compared to the standard LeNet architecture \cite{lecun1998gradient}. Architecture details are presented in Table-\ref{table:mnsit_architecure}.

\subsection{Impact of change in Hyperparameters}
\label{subsection:suppl_impact_of_hyperparameters}
\begin{figure}
\centering
        \fbox{\includegraphics[width=0.9\linewidth]{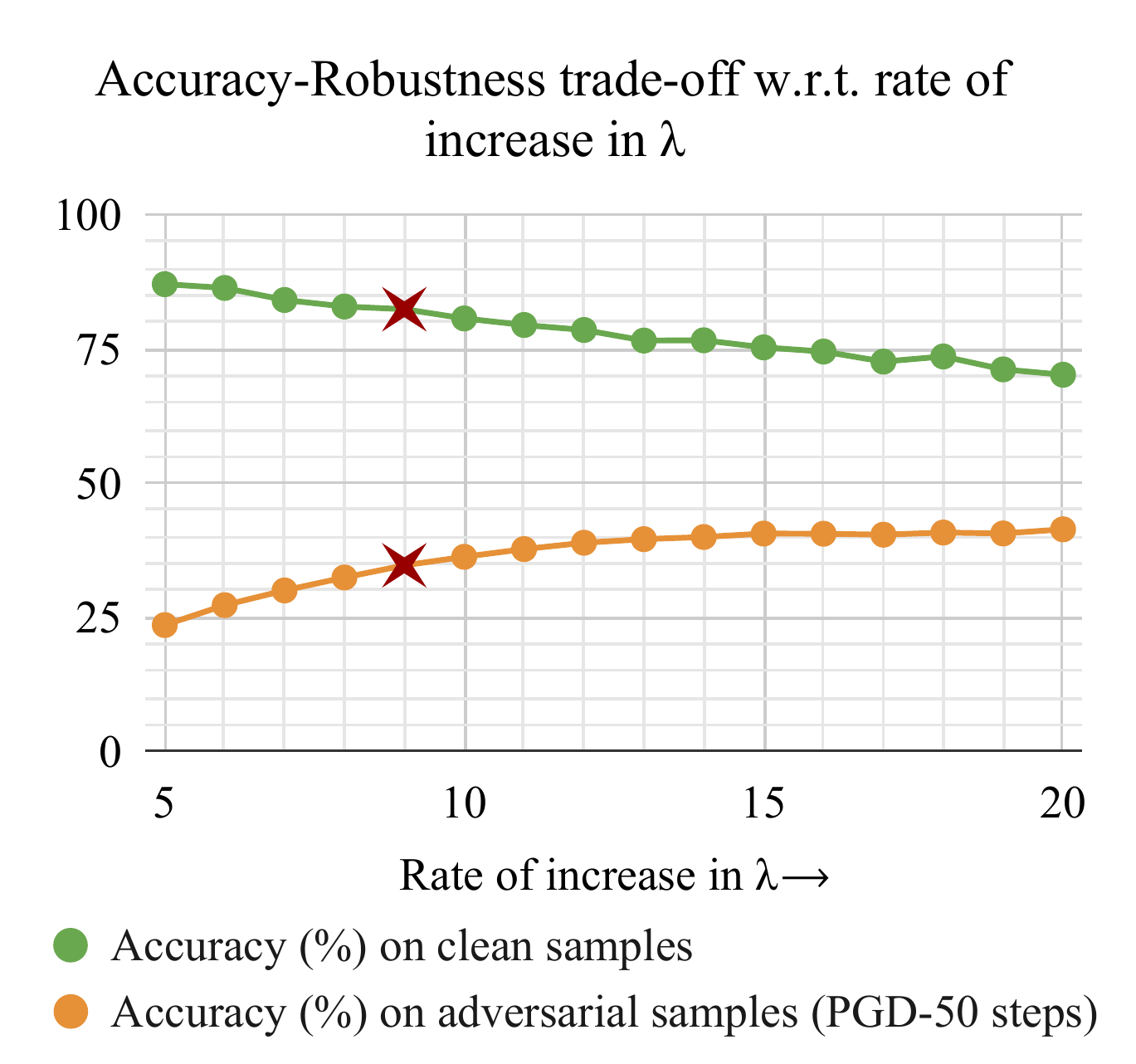}}
        \caption{Plot of  recognition accuracy (\%) on clean samples and PGD samples versus the rate of increase in hyperparameter($\lambda$) used for BPFC training. The selected setting (9) is highlighted using a cross mark.}
        \label{fig:acc_vs_hyp}
\vspace{-0.2cm}
\end{figure}

In this section, we study the effect of variation in the hyperparameter $\lambda$ (Eq.~(\ref{eq_loss}) in main paper). For CIFAR-$10$ dataset, we set the initial value of $\lambda$ to be $1$, and multiply this by a constant factor every $25$ epochs ($3$ times over $100$ epochs). We present the results obtained by changing the rate of increase in $\lambda$ for CIFAR-$10$ dataset in Fig-\ref{fig:acc_vs_hyp}. As the rate increases, accuracy on clean samples reduces, and accuracy on adversarial samples increases. The clean accuracy saturates to about $70\%$, and accuracy on adversarial samples saturates to approximately $40\%$. The best trade-off between both is obtained at a rate of 15, where the clean accuracy is $75.28\%$ and adversarial accuracy is $40.6\%$. However, for a fair comparison with PGD training and other existing methods, we select the rate at which clean accuracy matches with that of PGD-AT. Hence, the selected hyper-parameter is $9$. 

We use a similar methodology for hyperparameter selection in MNIST and Fashion-MNIST datasets as well. For these datasets, we set a fixed value of $\lambda$ and do not increase it over epochs. The value of $\lambda$ is selected such that the accuracy on clean samples matches with that of a PGD trained model.

\begin{table}[]
\caption{Network architectures used for Fashion-MNIST and MNIST datasets. Modified LeNet is used for training the model and Net-A is used as a source for generating black-box attacks.}
\centering
\label{table:mnsit_architecure}
\setlength\tabcolsep{8pt}{
    \begin{tabular}{@{}|c|c|@{}} 
    \toprule
    \textbf{Modified LeNet (M-LeNet)} & \textbf{Net-A} \\ \midrule
    \{conv(32,5,5) + Relu\}$\times$2 & Conv(64,5,5) + Relu \\
    MaxPool(2,2) & Conv(64,5,5) + Relu  \\
    \{conv(64,5,5) + Relu\}$\times$2 & Dropout(0.25)         \\
    MaxPool(2,2) & FC(128) + Relu  \\ 
    FC(512) + Relu & Dropout(0.5) \\
    FC + Softmax    & FC + Softmax \\
    \bottomrule
    \end{tabular}}
\vspace{0.2cm}
\end{table}

\section{Details on Experimental Results}
In this section, we present additional experimental results to augment our observations and results presented in the main paper.
\subsection{White-box attacks}
\label{subsection:suppl_whitebox_attacks}

\begin{table}[]
\caption{\textbf{Fashion-MNIST}: Recognition accuracy (\%) of models in a white-box attack setting.}
\label{table:fmnist_whitebox}
\vspace{-0.0cm}
\setlength\tabcolsep{3pt}
\resizebox{1.0\linewidth}{!}{
\begin{tabular}{l|ccc|ccc}
\toprule
\multirow{2}{*}{\textbf{\begin{tabular}[c]{@{}l@{}}Training \\ method\end{tabular}}}     & \multirow{2}{*}{\textbf{\begin{tabular}[c]{@{}l@{}}Clean\end{tabular}}} & \multirow{2}{*}{\textbf{\begin{tabular}[c]{@{}l@{}}FGSM\end{tabular}}} & \textbf{IFGSM} & \multicolumn{3}{c}{\textbf{PGD (n-steps)}}            \\
                     &                &               & \textbf{40 steps}        & \textbf{~~~~40~~~~} & \textbf{~~100~~} & \textbf{1000}                      \\\midrule
FGSM-AT               & 93.0 & 89.9 & 25.3 & 15.5 & 15.1 & 15.0 \\
RSS-AT                & 87.7 & 81.2 & 77.5 & 72.0 & 71.8 & 71.8 \\
PGD-AT                & 87.4 & 81.4 & 80.2 & 79.1 & 79.0 & 79.0 \\ \midrule
NT                & 92.0 & 16.6 & 2.4  & 0.3  & 0.3  & 0.3  \\
Mixup                 & 91.0 & 37.7 & 0.1  & 0.0  & 0.0  & 0.0     \\
BPFC (\textbf{Ours})  & 87.1 & 73.1 & 70.2 & 68.0 & 67.7 & 67.7 \\
 \bottomrule
\end{tabular}
}
\end{table}
\begin{table}[]
\caption{\textbf{MNIST}: Recognition accuracy (\%) of models in a white-box attack setting.}
\label{table:mnist_whitebox}
\vspace{-0.0cm}
\setlength\tabcolsep{3pt}
\resizebox{1.0\linewidth}{!}{
\begin{tabular}{l|ccc|ccc}
\toprule
\multirow{2}{*}{\textbf{\begin{tabular}[c]{@{}l@{}}Training \\ method\end{tabular}}}     & \multirow{2}{*}{\textbf{\begin{tabular}[c]{@{}l@{}}Clean\end{tabular}}} & \multirow{2}{*}{\textbf{\begin{tabular}[c]{@{}l@{}}FGSM\end{tabular}}} & \textbf{IFGSM} & \multicolumn{3}{c}{\textbf{PGD (n-steps)}}            \\
                     &                &               & \textbf{40 steps}        & \textbf{~~~~40~~~~} & \textbf{~~100~~} & \textbf{1000}                      \\\midrule

FGSM-AT               & 99.4 & 89.6 & 29.4 & 13.8 & 4.9  & 3.7  \\
RSS-AT                & 99.0 & 96.4 & 93.1 & 93.0 & 90.9 & 90.4 \\
PGD-AT                & 99.3 & 96.2 & 94.9 & 95.4 & 94.3 & 94.1 \\ \midrule
NT                & 99.2 & 82.7 & 0.5  & 0.0  & 0.0  & 0.0  \\
Mixup                 & 99.4 & 58.1 & 0.2  & 0.0  & 0.0  & 0.0  \\
BPFC (\textbf{Ours})  & 99.1 & 94.4 & 92.0 & 91.5 & 86.6 & 85.7 \\
 \bottomrule
\end{tabular}
}
\end{table}

\begin{table}[]
\caption{\textbf{PGD attack with multiple random restarts:} Recognition accuracy (\%) of different models on PGD adversarial samples with multiple random restarts in a white-box setting.}
\label{table:rr_all}
\setlength\tabcolsep{3pt}
\resizebox{1.0\linewidth}{!}{
\begin{tabular}{@{}l|ccc|ccc|ccc@{}}
\toprule
                \multirow{2}{*}{\textbf{\begin{tabular}[c|]{@{}l@{}}Training \\ method\end{tabular}}}& \multicolumn{3}{c|}{\textbf{CIFAR-10}} & \multicolumn{3}{|c|}{\textbf{Fashion-MNIST}} & \multicolumn{3}{c}{\textbf{MNIST}} \\ 
                & \multicolumn{3}{c|}{PGD 50-steps} & \multicolumn{3}{|c|}{PGD 100-steps} & \multicolumn{3}{c}{PGD 100-steps} \\
                \midrule
                \# restarts : & 1 & 100 & 1000 & 1 & 100 & 1000 & 1 & 100 & 1000   \\ \midrule
PGD-AT               & 45.3	& 44.9 & 44.9    & 80.6 & 79.7 & 79.6   & 92.9 & 90.9 & 90.6    \\
RSS-AT               & 45.2 & 44.7 & 44.7    & 74.1 & 73.4 & 73.2   & 88.7 & 86.3 & 85.6     \\
BPFC (\textbf{Ours}) & 35.6	& 35.1 & 34.9    & 69.4 & 68.4 & 68.3   & 84.0 & 80.5 & 79.9       \\ \bottomrule
\end{tabular}
}
\vspace{-0.1cm}
\end{table}

\subsubsection{Bounded attacks}
\label{subsection:suppl_bounded_attacks}
Detailed results on Fashion-MNIST and MNIST white-box attacks are presented in Tables-\ref{table:fmnist_whitebox} and \ref{table:mnist_whitebox} respectively. The proposed method achieves significantly better robustness to multi-step adversarial attacks when compared to Normal training (NT), FGSM-AT and Mixup. The robustness to multi-step attacks using the proposed approach is comparable to that of PGD-AT and RSS-AT models, while being faster than both approaches.

We run the PGD attack with multiple random restarts on a random sample of 1000 test set images, equally distributed across all classes. This experiment is done to ensure that the achieved robustness is not due to gradient masking. The results with random restarts are presented in Table-\ref{table:rr_all}. Here, the overall accuracy is computed as an average over the worst-case per-sample accuracy, as suggested by Carlini \etal \cite{carlini2019evaluating}. A $50$-step PGD attack is performed on CIFAR-$10$ dataset, and a $100$-step attack is performed on Fashion-MNIST and MNIST datasets. The degradation from $100$ random restarts to $1000$ random restarts is insignificant across all datasets, indicating the absence of gradient masking. Degradation from a single run to $100$ random restarts is also insignificant for CIFAR-$10$ and Fashion-MNIST. However, the degradation is larger for MNIST, similar to the trend observed with PGD-AT and RSS-AT models. It is to be noted that the results corresponding to this experiment may not coincide with those reported in Table-\ref{table:cifar10_whitebox} in the main paper, and Tables-\ref{table:fmnist_whitebox} and \ref{table:mnist_whitebox} in the Supplementary, as we consider only a sample of the test set for this experiment. 
\begin{table}[]
\caption{\textbf{DeepFool and C\&W attacks (Fashion-MNIST):} Average $\ell_2$ norm of the generated adversarial perturbations is reported. Higher $\ell_2$ norm implies better robustness. Fooling rate (FR) represents percentage of test set samples that are misclassified.}
\label{table:fmnist_deepfool_cw}
\vspace{-0.2cm}
\centering
\setlength\tabcolsep{3pt}
\resizebox{1.0\linewidth}{!}{
\begin{tabular}{@{}l|cc|cc@{}}
\toprule
\multirow{2}{*}{\textbf{\begin{tabular}[c]{@{}l@{}}Training \\ method\end{tabular}}} & \multicolumn{2}{|c}{\textbf{DeepFool}}                                   & \multicolumn{2}{|c}{\textbf{C\&W}}                                       \\ \cmidrule(l){2-5} 
                  & \multicolumn{1}{|c}{~~~FR $(\%)$~~~} & \multicolumn{1}{c|}{~~~Mean $\ell_2$~~~} & \multicolumn{1}{c}{~~~FR $(\%)$~~~} & \multicolumn{1}{c}{~~Mean $\ell_2$~~} \\ \midrule
FGSM-AT                 & 94.34 & 1.014 & 100.0 & 0.715   \\
PGD-AT                  & 90.70 & 3.429 & 100.0 & 2.142   \\
RSS-AT                  & 91.22 & 2.762 & 99.9 & 1.620   \\ \midrule
NT                      & 94.07 & 0.467 & 100.0 & 0.406   \\
Mixup                   & 92.22 & 0.226 & 100.0 & 0.186   \\
BPFC (\textbf{Ours})    & 90.94 & 3.620 & 100.0 & 1.789 \\ \bottomrule
\end{tabular}
}
\end{table}

\begin{table}[]
\caption{\textbf{DeepFool and C\&W attacks (MNIST):} Average $\ell_2$ norm of the generated adversarial perturbations is reported. Higher $\ell_2$ norm implies better robustness. Fooling rate (FR) represents percentage of test set samples that are misclassified.}
\label{table:mnist_deepfool_cw}
\vspace{-0.2cm}
\centering
\setlength\tabcolsep{3pt}
\resizebox{1.0\linewidth}{!}{
\begin{tabular}{@{}l|cc|cc@{}}
\toprule
\multirow{2}{*}{\textbf{\begin{tabular}[c]{@{}l@{}}Training \\ method\end{tabular}}} & \multicolumn{2}{|c}{\textbf{DeepFool}}                                   & \multicolumn{2}{|c}{\textbf{C\&W}}                                       \\ \cmidrule(l){2-5} 
                  & \multicolumn{1}{|c}{~~~FR $(\%)$~~~} & \multicolumn{1}{c|}{~~~Mean $\ell_2$~~~} & \multicolumn{1}{c}{~~~FR $(\%)$~~~} & \multicolumn{1}{c}{~~Mean $\ell_2$~~} \\ \midrule
FGSM-AT                 & 99.36  & 3.120 & 100.0 & 1.862   \\
PGD-AT                  & 95.97 & 5.316 & 100.0 & 3.053   \\
RSS-AT                  & 94.41 & 4.894 & 98.2 & 2.725   \\ \midrule
NT                      & 99.15 & 1.601 & 100.0 & 1.427   \\
Mixup                   & 91.82 & 0.518 & 100.0 & 0.498   \\
BPFC (\textbf{Ours})    & 97.47 & 6.289 & 100.0 & 3.041  \\ \bottomrule
\end{tabular}
}
\vspace{-0.2cm}
\end{table}

\subsubsection{Unbounded attacks}
\label{subsection:suppl_unbounded_attacks}
The results with unbounded attacks (DeepFool \cite{moosavi2016deepfool} and Carlini-Wagner (C\&W) \cite{carlini2017towards}) for Fashion-MNIST and MNIST datasets are presented in Tables-\ref{table:fmnist_deepfool_cw} and \ref{table:mnist_deepfool_cw} respectively. We select the following hyperparameters for C\&W attack on Fashion-MNIST and MNIST datasets: search steps = $9$, max iterations = $500$, learning rate = $0.01$. For DeepFool attack, we set the number of steps to $100$ for both Fashion-MNIST and MNIST datasets.

The average $\ell_2$-norm of the generated perturbations to achieve approximately $100\%$ fooling rate using C\&W attack is higher with the proposed approach when compared to most other approaches, with the exception of PGD-AT, whose average $\ell_2$-norm is marginally higher. DeepFool attack does not achieve $100\%$ fooling rate for Fashion-MNIST and MNIST datasets, as was the case with CIFAR-$10$ (ref: Section-\ref{subsection:experiments_wb_unbounded} of main paper). However, since the fooling rates of the proposed approach are comparable to, or greater than that of PGD-AT and RSS-AT, we can make a fair comparison between the required $\ell_2$-norm for achieving the given fooling rate across these approaches. We observe that the proposed approach is more robust to DeepFool attack, when compared to both of these approaches. 

\subsection{Black-box attacks}
\label{subsection:suppl_blackbox_attacks}
Multi-step attacks such as I-FGSM are known to show weak transferability across models in a black-box setting \cite{kurakin2016adversarial}. Dong~\etal\cite{dong2018boosting} introduced a momentum term in the optimization process of I-FGSM, so as to increase the transferability of the generated adversarial samples. This attack is referred to as the Momentum Iterative FGSM (MI-FGSM) attack. 

The results corresponding to black-box multi-step PGD and MI-FGSM attacks are presented in Tables-\ref{table:pgd_bb} and \ref{table:mifgsm_bb} respectively. We consider two source models for black-box attacks on each of the models trained: one with the same architecture as the target model, and second with a different architecture. For Fashion-MNIST and MNIST, the architecture of the second model (Net-A) is presented in Table-\ref{table:mnsit_architecure}. For CIFAR-$10$, we consider a second model with VGG-19 \cite{simonyan2014very} architecture. 
The proposed approach achieves a significant improvement in robustness to adversarial samples with respect to Normal Training (NT) and Mixup, and comparable results with respect to the adversarial training methods, across all the datasets. 

\begin{table}[]
\caption{\textbf{PGD Black-box attacks:} Recognition accuracy (\%) of different models on PGD black-box adversaries. Columns represent source model used for generating the attack. $7$-step attack is used for CIFAR-$10$ and $40$-step attack is used for Fashion-MNIST and MNIST}
\label{table:pgd_bb}
\vspace{-0.2cm}
\setlength\tabcolsep{3pt}
\resizebox{1.0\linewidth}{!}{
\begin{tabular}{@{}l|cc|cc|cc@{}}
\toprule
                \multirow{2}{*}{\textbf{\begin{tabular}[c]{@{}l@{}}Training \\ method\end{tabular}}}& \multicolumn{2}{|c|}{\textbf{CIFAR-10}} & \multicolumn{2}{|c|}{\textbf{Fashion-MNIST}} & \multicolumn{2}{c}{\textbf{MNIST}} \\ \cmidrule(l){2-7} 
                & VGG19      & ResNet18      & Net-A           & M-LeNet       & Net-A           & M-LeNet      \\
                \midrule
FGSM-AT         & 85.85 & 85.61 & 94.27 & 91.52 & 79.8 & 74.11      \\
RSS-AT          & 80.92 & 80.82 & 84.71 & 83.91 & 95.19 & 96.27      \\
PGD-AT          & 81.37 & 81.22 & 85.16 & 85.71 & 96.52 & 96.69    \\\midrule
NT              & 16.86 & 0 & 27.10 & 0.33 & 4.64 & 0.03      \\
Mixup           & 30.16 & 29.53 & 49.07 & 60.71 & 31.4 & 58.25      \\
BPFC (\textbf{Ours}) & 80.42 & 80.15 & 81.45 & 83.00 & 95.31 & 95.91   \\ \bottomrule
\end{tabular}
}
\end{table}
\begin{table}[]
\caption{\textbf{MI-FGSM \cite{dong2018boosting} Black-box attacks:} Recognition accuracy (\%) of different models on MI-FGSM black-box adversaries. Columns represent source model used for generating the attack. $7$-step attack is used for CIFAR-$10$ and $40$-step attack is used for Fashion-MNIST and MNIST}
\label{table:mifgsm_bb}
\vspace{-0.2cm}
\setlength\tabcolsep{3pt}
\resizebox{1.0\linewidth}{!}{
\begin{tabular}{@{}l|cc|cc|cc@{}}
\toprule
                \multirow{2}{*}{\textbf{\begin{tabular}[c]{@{}l@{}}Training \\ method\end{tabular}}}& \multicolumn{2}{|c|}{\textbf{CIFAR-10}} & \multicolumn{2}{|c|}{\textbf{Fashion-MNIST}} & \multicolumn{2}{c}{\textbf{MNIST}} \\ \cmidrule(l){2-7} 
                & VGG19      & ResNet18      & Net-A           & M-LeNet       & Net-A           & M-LeNet      \\
                \midrule
FGSM-AT         & 76.44 & 74.22 & 94.61 & 92.11 & 79.95 & 73.92      \\
RSS-AT          & 80.21 & 80.10 & 84.61 & 84.02 & 96.11 & 95.28       \\
PGD-AT          & 80.47 & 80.59 & 84.98 & 85.58 & 95.56 & 95.34      \\\midrule
NT              & 12.98 & 0.04 & 28.28 & 4.69 & 12.48 & 1.93         \\
Mixup           & 35.74 & 25.22 & 50.60 & 63.32 & 43.72 & 62.98        \\
BPFC (\textbf{Ours}) & 79.04 & 79.04 & 81.33 & 82.70 & 94.03 & 94.46   \\ \bottomrule
\end{tabular}
}
\vspace{-0.2cm}
\end{table}

\subsection{Adaptive attacks}
\label{subsection:suppl_adaptive_attacks}
In this section, we explain the adaptive attacks used in this paper in greater detail. We utilize information related to the proposed regularizer to construct potentially stronger attacks when compared to a standard PGD attack. 
We maximize the following loss function to generate an adaptive attack corresponding to each data sample $x_i$:

\vspace{-0.5cm}

\begin{multline}
\label{eq:adaptive_loss}
       L_{i} = \lambda_{ce}~ce(f(x_i),y_i) + \lambda_{g} {\norm{g(x_i) - g(q(x_i))}}^2_2 \\
    - \lambda_{LSB} {\norm{x_i - q(x_i)}}^2_2 
\end{multline}

The quantized image corresponding to $x_i$ is denoted by $q(x_i)$. We consider $f(.)$ as the function mapping of the trained network, from an image $x_i$, to its corresponding softmax output $f(x_i)$. The corresponding pre-softmax output of the network is denoted by $g(x_i)$. The ground truth label corresponding to $x_i$ is denoted by $y_i$. The first term in the above equation is the cross-entropy loss, the second term is the BPFC regularizer proposed in this paper, and the third term is an $\ell_2$ penalty term on the magnitude of $k$ LSBs. We consider the value of $k$ to be the same as that used for training the models (ref: Section-\ref{subsection:experiments_Preliminaries} in the main paper). The coefficients of these loss terms are denoted by $\lambda_{ce}$, $\lambda_{g}$ and $\lambda_{LSB}$ respectively.

Maximizing the cross-entropy term leads to finding samples that are misclassified by the network. Maximizing the BPFC loss results in finding samples which do not comply with the BPFC regularizer imposed during training. Minimizing the third term would help find samples with low magnitude LSBs, which are possibly the points where the defense is less effective. The objective of an adversary is to cause misclassification, which can be achieved by maximizing only the first term in Eq.~(\ref{eq:adaptive_loss}). However, the proposed defense mechanism could lead to masking of the true solution, thereby resulting in a weak attack. Thus, the role of the remaining terms, which take into account the defense mechanism, is to aid the optimization process in finding such points, if any. The remainder of the algorithm used is similar to that proposed by Madry \etal \cite{madry-iclr-2018}. 
\begin{table}[]
\caption{\textbf{CIFAR-10}: Recognition accuracy (\%) of the model trained using the proposed approach on adversarial samples generated using adaptive attacks.}
\label{table:cifar10_adaptive}
\vspace{-0.0cm}
\setlength\tabcolsep{3pt}
\resizebox{1.0\linewidth}{!}{
\begin{tabular}{l|ccc|c|c|c}
\toprule
\multirow{2}{*}{\textbf{\begin{tabular}[|c|]{@{}l@{}}Adaptive attack\end{tabular}}}& \multicolumn{3}{c}{\textbf{Loss coefficients}}                                   & \multicolumn{3}{|c}{\textbf{n-step Adaptive attack}}                                       \\ \cmidrule(l){2-7} 

    & ~~$\lambda_{ce}~~$ & ~$\lambda_{g}$~ & $\lambda_{LSB}$ & ~~~~~7~~~~~  & ~~~~20~~~~  & ~~~~50~~~~ 
\\\midrule\midrule
PGD               & 1 & 0 & 0 & 41.72 & 35.74 & 34.68 \\ \midrule
\multirow{3}{*}{\begin{tabular}[c]{@{}l@{}}Variation in $\lambda_{g}$\\ ($\lambda_{ce}$ = 1 and \\ $\lambda_{LSB}$ = 0)\end{tabular}}  
                & 1	& 0.1 &	0 &	41.67 &	35.65 &	34.61  \\
                & 1	& 1	& 0	& 41.49	& 35.42	& 34.52  \\
                & 1	& 10 & 0 & 42.15 & 36.14 & 35.30 \\ \midrule
\multirow{4}{*}{\begin{tabular}[c]{@{}l@{}}Variation in $\lambda_{g}$\\ ($\lambda_{ce}$ = 0 and \\ $\lambda_{LSB}$ = 0)\end{tabular}} 
 & 0 & 0.1 & 0 & 41.65 & 35.62 & 34.58 \\
 & 0 & 0.5 & 0 & 41.54 & 35.45 & 34.41 \\
 & 0 &	1  & 0 & 64.35 & 59.95 & 59.16 \\
 & 0 & 10 & 0 &	42.15 &	36.15 & 35.30 \\ \midrule
\multirow{2}{*}{\begin{tabular}[c]{@{}l@{}}Variation in $\lambda_{LSB}$\\ ($\lambda_{ce}$ = 1 and \\ $\lambda_{g}$ = 0)\end{tabular}}&&&&&& \\ 
& 1 & 0 & 1 & 42.00 & 35.96 & 34.89 \\
& 1 & 0 & 10 & 48.49 & 41.40 & 39.60 \\ \midrule
\multirow{2}{*}{\begin{tabular}[c]{@{}l@{}}Variation in $\lambda_{LSB}$\\ ($\lambda_{ce}$ = 1 and \\ $\lambda_{g}$ = 1)\end{tabular}}&&&&&& \\ 
& 1 & 1 & 1 & 41.67 & 35.47 & 34.52 \\
& 1 & 1 & 10 & 46.07 & 37.54 & 35.79 \\ \bottomrule
\end{tabular}
}
\end{table}

The results with adaptive attacks for CIFAR-$10$, Fashion-MNIST and MNIST datasets are presented in Tables-\ref{table:cifar10_adaptive}, \ref{table:fmnist_adaptive} and \ref{table:mnist_adaptive} respectively.  We consider the following coefficients in Eq.~(\ref{eq:adaptive_loss}) to find a strong adaptive attack: 
\begin{itemize}
\itemsep0em 
    \item $\lambda_{ce}=1,~ \lambda_{g}=0,~ \lambda_{LSB}=0$ \\
    This corresponds to a standard PGD attack \cite{madry-iclr-2018}, which serves as a baseline in this table. The goal of the remaining experiments is to find an attack stronger than this. 
    \item $\lambda_{ce}=1,~ \lambda_{g}=$ variable, $\lambda_{LSB}=0$ \\
    This case corresponds to using the training loss directly to find adversarial samples. We find that lower values of $\lambda_{g}$ lead to stronger attacks, while still not being significantly stronger than baseline. This indicates that addition of the BPFC regularizer does not help in the generation of a stronger attack.
    \item $\lambda_{ce}=0,~ \lambda_{g}=$ variable, $\lambda_{LSB}=0$ \\
    For CIFAR-$10$ dataset, this case is able to generate attacks which are as strong as PGD, without using the cross-entropy term. This indicates that the BPFC loss term is relevant in the context of generating adversarial samples. However, addition of this to the cross-entropy term does not generate a stronger attack, as the defense is not masking gradients that prevents generation of stronger adversaries. However, for Fashion-MNIST and MNIST datasets, this attack is weaker than PGD.
    \item $\lambda_{ce}=1,~ \lambda_{g}=$ variable, $\lambda_{LSB}=$ variable \\
    Next, we consider the case of introducing the third term that imposes a penalty on high magnitude LSBs. Addition of this term with or without the BPFC term does not help generate a stronger attack, indicating that this training regime does not create isolated points in the $\ell_{\infty}$-ball around each sample, which correspond to points with low magnitude LSBs. This can be attributed to the addition of pre-quantization noise. 
\end{itemize}

Overall, the adaptive attacks constructed based on the knowledge of the defense mechanism do not lead to stronger attacks. This leads to the conclusion that the proposed defense does not merely make the process of finding adversaries harder, but results in learning models that are truly robust.

\begin{table}[]
\caption{\textbf{Fashion-MNIST}: Recognition accuracy (\%) of the model trained using the proposed approach on adversarial samples generated using adaptive attacks.}
\label{table:fmnist_adaptive}
\vspace{-0.0cm}
\setlength\tabcolsep{3pt}
\resizebox{1.0\linewidth}{!}{
\begin{tabular}{l|ccc|c|c|c}
\toprule
\multirow{2}{*}{\textbf{\begin{tabular}[|c|]{@{}l@{}}Adaptive attack\end{tabular}}}& \multicolumn{3}{c}{\textbf{Loss coefficients}}                                   & \multicolumn{3}{|c}{\textbf{n-step Adaptive attack}}                                       \\ \cmidrule(l){2-7} 

    & ~~$\lambda_{ce}~$ & ~$\lambda_{g}$~ & $\lambda_{LSB}$ & ~~~~40~~~~  & ~~~100~~~  & ~~~500~~~ 
\\\midrule\midrule
PGD            & 1 & 0 & 0 & 68.03 & 67.75 & 67.71    \\ \midrule
\multirow{3}{*}{\begin{tabular}[c]{@{}l@{}}Variation in $\lambda_{g}$\\ ($\lambda_{ce}$ = 1 and \\ $\lambda_{LSB}$ = 0)\end{tabular}}  
            & 1 & 1  & 0 & 69.41 & 69.22 & 69.19 \\
            & 1 & 10 & 0 & 76.72 & 76.44 & 76.46 \\
            & 1 & 25 & 0 & 78.95 & 78.8  & 78.79    \\ \midrule
\multirow{4}{*}{\begin{tabular}[c]{@{}l@{}}Variation in $\lambda_{g}$\\ ($\lambda_{ce}$ = 0 and \\ $\lambda_{LSB}$ = 0)\end{tabular}} 
            & 0 & 1  & 0 & 80.45 & 80.23 & 80.2  \\
            & 0 & 10 & 0 & 80.46 & 80.24 & 80.22 \\
            & 0 & 25 & 0 & 80.46 & 80.22 & 80.18 \\
            & 0 & 50 & 0 & 80.46 & 80.22 & 80.18\\ \midrule
\multirow{2}{*}{\begin{tabular}[c]{@{}l@{}}Variation in $\lambda_{LSB}$\\ ($\lambda_{ce}$ = 1 and \\ $\lambda_{g}$ = 0)\end{tabular}}&&&&&& \\ 
  & 1 & 0 & 1  & 68.2  & 67.98 & 67.95 \\
 & 1 & 0 & 10 & 71.32 & 70.98 & 70.98  \\ \midrule
\multirow{2}{*}{\begin{tabular}[c]{@{}l@{}}Variation in $\lambda_{LSB}$\\ ($\lambda_{ce}$ = 1 and \\ $\lambda_{g}$ = 25)\end{tabular}}&&&&&& \\ 
   & 1 & 25 & 1  & 78.96 & 78.8  & 78.77 \\
   & 1 & 25 & 10 & 78.91 & 78.74 & 78.76  \\ \bottomrule
\end{tabular}
}
\vspace{0.1cm}
\end{table}
\begin{table}[]
\caption{\textbf{MNIST}: Recognition accuracy (\%) of the model trained using the proposed approach on adversarial samples generated using adaptive attacks.}
\label{table:mnist_adaptive}
\vspace{-0.0cm}
\setlength\tabcolsep{3pt}
\resizebox{1.0\linewidth}{!}{
\begin{tabular}{l|ccc|c|c|c}
\toprule
\multirow{2}{*}{\textbf{\begin{tabular}[|c|]{@{}l@{}}Adaptive attack\end{tabular}}}& \multicolumn{3}{c}{\textbf{Loss coefficients}}                                   & \multicolumn{3}{|c}{\textbf{n-step Adaptive attack}}                                       \\ \cmidrule(l){2-7} 

    & ~~$\lambda_{ce}~$ & ~$\lambda_{g}$~ & $\lambda_{LSB}$ & ~~~~40~~~~  & ~~~100~~~  & ~~~500~~~ 
\\\midrule\midrule
PGD               & 1 & 0 & 0 & 91.49 & 86.6 & 85.63 \\ \midrule
\multirow{3}{*}{\begin{tabular}[c]{@{}l@{}}Variation in $\lambda_{g}$\\ ($\lambda_{ce}$ = 1 and \\ $\lambda_{LSB}$ = 0)\end{tabular}}  
               & 1 & 1  & 0 & 92.99 & 89.13 & 88.2  \\
               & 1 & 10 & 0 & 94.58 & 91.75 & 91.04 \\
               & 1 & 30 & 0 & 94.74 & 91.99 & 91.26 \\ \midrule
\multirow{4}{*}{\begin{tabular}[c]{@{}l@{}}Variation in $\lambda_{g}$\\ ($\lambda_{ce}$ = 0 and \\ $\lambda_{LSB}$ = 0)\end{tabular}} 
 & 0 & 1  & 0 & 94.8  & 91.96 & 91.34 \\
 & 0 & 10 & 0 & 94.8  & 91.97 & 91.34 \\
 & 0 & 30 & 0 & 94.79 & 91.98 & 91.38 \\
 & 0 & 50 & 0 & 94.79 & 91.97 & 91.35 \\ \midrule
\multirow{2}{*}{\begin{tabular}[c]{@{}l@{}}Variation in $\lambda_{LSB}$\\ ($\lambda_{ce}$ = 1 and \\ $\lambda_{g}$ = 0)\end{tabular}}&&&&&& \\ 
 & 1 & 0 & 1  & 91.56 & 86.96 & 86.01 \\
 & 1 & 0 & 10 & 93.51 & 90.11 & 89.41 \\ \midrule
\multirow{2}{*}{\begin{tabular}[c]{@{}l@{}}Variation in $\lambda_{LSB}$\\ ($\lambda_{ce}$ = 1 and \\ $\lambda_{g}$ = 30)\end{tabular}}&&&&&& \\ 
 & 1 & 30 & 1  & 94.72 & 91.98 & 91.33 \\
 & 1 & 30 & 10 & 94.7  & 91.97 & 91.36  \\ \bottomrule
\end{tabular}
}
\end{table}
\begin{figure*}
\centering
        \includegraphics[width=\linewidth]{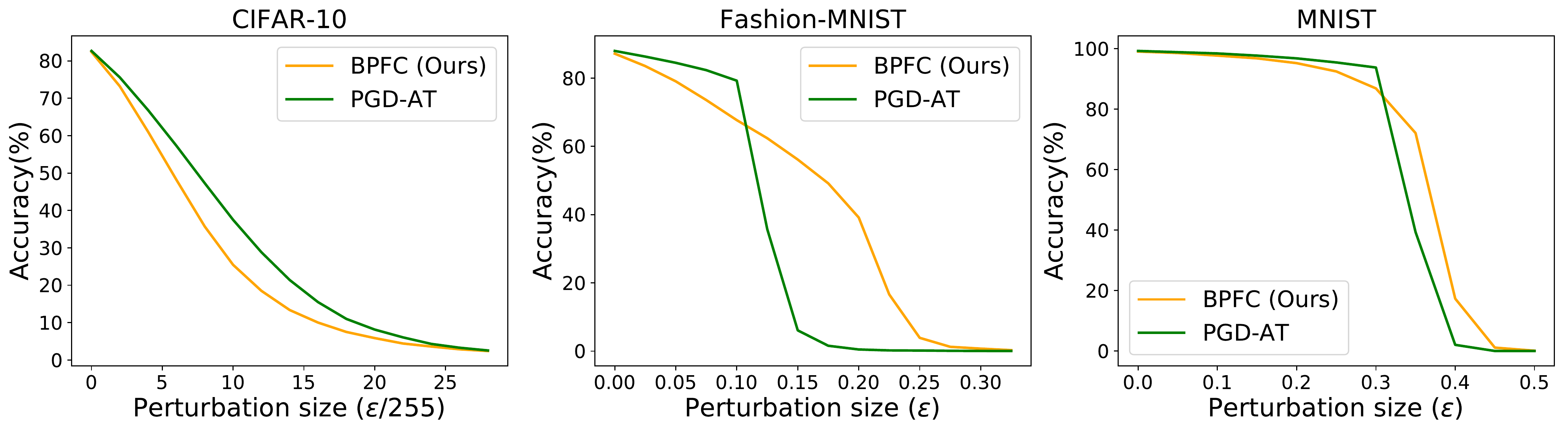}
        \vspace{-0.6cm}
        \caption{Plot of  recognition accuracy (\%) on PGD samples generated on test set versus perturbation size of PGD $7$-step attack. The model's accuracy is zero for large perturbation sizes indicating the absence of gradient masking.}
        \label{fig:acc_vs_eps}
        \vspace{-0.2cm}
\end{figure*}

\begin{figure*}
        \includegraphics[width=\linewidth]{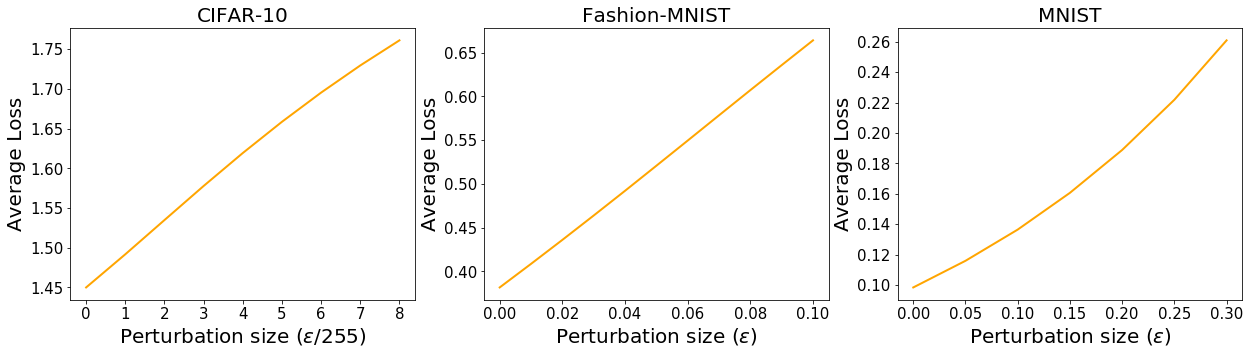}
        \caption{Plot of average loss on FGSM samples generated on test set versus perturbation size of FGSM attack.}
        \label{fig:loss_vs_eps}
        \vspace{-0.2cm}
\end{figure*}

\begin{table}[]
\caption{\textbf{ImageNet (white-box attacks):} Recognition accuracy (\%) of BPFC trained model and PGD-AT model on ImageNet dataset under white-box attack setting. Accuracy is reported on the following untargeted attacks: FGSM attack, I-FGSM $20$-step attack (IFGSM), PGD $20$-step attack and PGD $100$-step attack. Accuracy on PGD $20$-step targeted attack (with random targets) is also reported (TPGD).}
\label{table:imagenet}
\vspace{-0.2cm}
\setlength\tabcolsep{3pt}
\resizebox{1.0\linewidth}{!}{
\begin{tabular}{@{}l|ccccccc@{}}
\toprule
            \multirow{2}{*}{\textbf{\begin{tabular}[c]{@{}l@{}}Training \\method\end{tabular}}} & \multirow{2}{*}{\textbf{\begin{tabular}[c]{@{}l@{}}Clean\end{tabular}}} & \multirow{2}{*}{\textbf{\begin{tabular}[c]{@{}l@{}}FGSM\end{tabular}}}  & \textbf{IFGSM} & \textbf{PGD} & \textbf{PGD} & \textbf{TPGD}  \\ 
            
           &  &   & (20)& (20) & (100) & (20) \\\midrule
PGD-AT      & 47.91 & 24.42 & 21.52    & 19.39  & 19.06   & 43.43                     \\
BPFC (Ours) & 40.82 & 19.97 & 15.93    & 13.41  & 12.82   & 32.91                \\\bottomrule
\end{tabular}}
\vspace{-0.1cm}
\end{table}

\subsection{Basic Sanity Checks to verify Robustness}
\label{subsection:suppl_sanity_checks}
In this section, we present details related to Section-\ref{subsection:sanity_checks} in the main paper. The plots of accuracy verses perturbation size in Fig.\ref{fig:acc_vs_eps} demonstrate that unbounded attacks are able to reach $100\%$ success rate. It can be observed that increasing the distortion bound increases the success rate of the attack. Fig-\ref{fig:loss_vs_eps} shows a plot of the average loss on FGSM samples generated on the test set, versus perturbation size of the FGSM attack. It can be observed that the loss increases monotonically with an increase in perturbation size. These two plots confirm that there is no gradient masking effect \cite{athalye2018obfuscated} in the models trained using the proposed approach.

\subsection{Results on ImageNet}
\label{subsection:suppl_imagenet}
We report results on ImageNet dataset using the proposed method and PGD-AT in Table-\ref{table:imagenet}. The architecture used for both methods is ResNet-$50$ \cite{he2016deep}. We use the PGD-AT pre-trained model from \cite{robustness} for comparison. We train the proposed method for $125$ epochs and decay learning rate by a factor of $10$ at epochs $35$, $70$ and $95$. Similar to CIFAR-$10$, we start with a $\lambda$ of $1$ and step it up by a factor of $9$ at epochs $35$ and $70$. We use a higher step-up factor of $20$ at epoch $95$ to improve robustness. Since training ImageNet models is computationally intensive, we report results using similar hyperparameters as that of CIFAR-$10$. However, tuning hyperparameters specifically for ImageNet can lead to improved results. Accuracy on black-box FGSM attack is $47.39\%$ for PGD-AT and $40.41\%$ for the BPFC trained model. We note that the trend in robustness when compared to PGD-AT is similar to that of CIFAR-$10$, thereby demonstrating the scalability of the proposed approach to large scale datasets.

\end{document}